\documentclass{article}

\usepackage[preprint]{neurips_2026}
\usepackage[utf8]{inputenc} 
\usepackage[T1]{fontenc}    
\usepackage{hyperref}       
\usepackage{url}            
\usepackage{booktabs}       
\usepackage{amsfonts}       
\usepackage{nicefrac}       
\usepackage{microtype}      
\usepackage{xcolor}         
\usepackage{amsmath} 
\usepackage{graphicx}
\usepackage[table]{xcolor}
\usepackage{float} 
  \usepackage{subcaption}
  \usepackage{multirow}
\title{Rethinking Handwritten Character Recognition}

\author{%
  Ranjit Raut \\
  Department of Artificial Intelligence\\
  Kathmandu University\\
  Dhulikhel, Nepal \\
  \And
  Aarav Subedi \\
  Department of Artificial Intelligence\\
  Kathmandu University\\
  Dhulikhel, Nepal \\
  \And
  Ashim Shrestha \\
  Department of Artificial Intelligence\\
  Kathmandu University\\
  Dhulikhel, Nepal \\
}

\begin{document}

\maketitle

\begin{abstract}
Non-Latin handwritten character recognition (HCR) remains understudied. Dominant methods consider it as generic image classification, which uses model scale to implicitly learn stroke structure. Structural-prior efficiency---the principle that explicitly encoding script-geometric regularities as architectural inductive biases can be both more accurate and require fewer parameters. We introduce GraphemeNet, a unified multi-script architecture, governed by two orthogonal binary axes. Axis 1 operationalises stroke-level geometric regularity via Persistent Scaffold Injection (PSI): a script-specific asymmetric convolution injects a stroke scaffold as a weighted residual at every encoder stage, continuously anchoring learned features to script geometry---distinct from skip connections, auxiliary losses, or attention reweighting. Axis 2 selects between global average pooling with gated fusion and cross-scale attention with a Stroke Topology Module (STM), depending on whether glyph discrimination requires spatial relational reasoning. A Linear Capsule Routing (LCR) with $O(n)$ routing is shared universally. On fourteen benchmarks across eight writing systems, the architecture generalises with only scaffold and decoder topology varying per script, consistently challenging, outperforming published baselines, and establishing structural-prior efficiency as a broadly applicable principle for multi-script HCR.
\end{abstract}

\section{Introduction}
\label{sec:intro}
 
Handwritten character recognition (HCR) has achieved near-human accuracy on
Latin benchmarks~\cite{lecun1998,cohen2017emnist}, yet this success conceals a
fundamental inefficiency: dominant architectures compensate for the absence of
structural knowledge by scaling parameters.
A ResNet-85 requires ${\sim}39$\,M parameters to reach 99.72\% on the
Devanagari DHCD benchmark~\cite{mishra2021}; MallaNet~\cite{malla2025}
achieves 99.71\% at 17\,M parameters through a branching CNN with homogeneous
filter capsules.
Both improvements represent incremental rescaling rather than a change in the
underlying representational strategy.
The question they raise is not whether scale works, but whether it is
\emph{necessary}—and whether the parameter burden reflects a solvable
architectural deficit rather than an irreducible property of the task.
 
This deficit becomes acute when recognition extends across writing systems.
Every script encodes meaning in stroke geometry: Devanagari and Bengali
characters hang from the \emph{shirorekha}, a horizontal top-bar whose
continuity and position carry class-discriminative weight.
Arabic is a right-to-left cursive abjad in which letter identity depends
jointly on base-stroke shape and the count and position of dot diacritics.
Thai consonants are defined by circular loops at varying heights above the
baseline.
Kuzushiji cursive script carries flowing diagonal strokes with explicit
harai tails whose angular variation encodes character class.
Each of these regularities is consistent enough across writers to constitute
a reliable geometric prior—yet no prior HCR architecture encodes these priors
\emph{persistently} across the encoder hierarchy.
Existing work inserts structural signals at isolated points: as auxiliary
preprocessing~\cite{jayadevan2011,singh2023}, single-stage attention
reweighting~\cite{hu2018,woo2018}, or stem-only asymmetric
convolutions~\cite{szegedy2016}.
Once the signal passes the first encoder stage it competes with, and is
progressively diluted by, learned texture features.
The stroke detail that distinguishes a Devanagari \textit{ka} from
\textit{kha}—a single stroke at a specific spatial position—has no guaranteed
anchor in the network's representational hierarchy.
 
We address this deficit with two orthogonal design axes, each grounded in a
measurable property of the target script.
\textbf{Axis~1} asks whether the script possesses consistent stroke-level
geometric regularity amenable to persistent structural anchoring.
\textbf{Axis~2} asks whether glyph discrimination requires reasoning about
spatial relationships between strokes \emph{across encoder scales}, or whether
local post-anchoring features are sufficient.
These two binary decisions partition the space of writing systems into
\textbf{four empirically validated cells} and determine the optimal
configuration of GraphemeNet for any script. To operationalise Axis~1, we introduce \textbf{Persistent Scaffold Injection
(PSI)}: a script-specific asymmetric convolution extracts a stroke scaffold from
the raw input, and this scaffold is spatially downsampled, channel-projected, and
injected as a weighted residual ($\alpha = 0.1$) at \emph{every} encoder stage.
PSI is not a skip connection, which carries full feature maps between encoder
stages; not an auxiliary loss, which operates at a single point; and not an
attention mechanism, which reweights existing features rather than introducing
an external structural signal.
It is the only component that varies across scripts — a single kernel-shape
parameter (Table~\ref{tab:scaffold}).
To operationalise Axis~2, we provide two interchangeable decoder modules: a
multi-scale global average pooling decoder with 2-way gated fusion for scripts
where structural priors suffice, and a cross-scale attention module with a
Stroke Topology Module (STM) and 3-way gated fusion for scripts requiring
spatial relational reasoning.
A \textbf{Linear Capsule Routing (LCR)} providing $\mathcal{O}(n)$
class-discriminative routing is shared universally across all variants as a
complement to both decoder paths.

The axis assignments are predictive, not post-hoc.
Indic and Semitic scripts (Devanagari, Bengali, Kannada, Arabic, Persian)
achieve peak accuracy under PSI with the GAP decoder: the scaffold prior is
sufficient for local feature discrimination, and cross-scale attention adds
no complementary information.
Japanese Kuzushiji requires PSI combined with cross-scale attention between Option~B and Option~A
constitutes the strongest evidence in this paper, directly demonstrating that
cursive scripts with spatially complex glyph topology require relational spatial
decoding that global average pooling cannot provide.
Thai and English require cross-scale attention without PSI: their complex
loop geometry and case-sensitive ambiguity both need relational decoding, but
a scalar scaffold bypass would smear rather than anchor their geometric
structure.
Notably, \emph{no script} in our study is best served by the no-PSI + GAP
decoder combination, confirming that these two axes are jointly and
predictively sufficient.

\section{Related Work}
\label{sec:related}
Existing multi-script recognition work follows one of two strategies:
building independent script-specific models, or training a shared feature
extractor with script-specific classification heads.
Comparative studies such as Parvez et al.~\cite{parvezreview2025} evaluate
multiple architectures across scripts but do not propose a unified architecture
designed for multi-script generalisation.
Multi-task learning approaches treat script identity as an auxiliary
task~\cite{multiscriptmtl2021}, improving calibration but not architecture.
Transformer-based systems such as TrOCR~\cite{li2023trocr} fine-tune a shared
backbone per language, treating script structure as implicit in training data
rather than encoding it architecturally.
No published work, to our knowledge, evaluates a single architectural
skeleton—with script-local variation confined to a single structural
parameter—across eight typologically distinct writing systems covering Brahmic,
Semitic, East Asian, Southeast Asian, and Latin scripts simultaneously. 

The dominant paradigm in handwritten character recognition treats isolated
character images as generic visual objects and relies on model scale to
implicitly learn stroke structure.
On the Devanagari DHCD benchmark, Mishra et al.\ achieved 99.72\% with a
ResNet-85 model of ${\sim}39$\,M parameters~\cite{mishra2021};
MallaNet subsequently improved parameter efficiency to 17\,M through a
branch-merge CNN with homogeneous filter capsules~\cite{malla2025}, though
both results remain scale-driven rather than structure-driven.
For Bengali, Shahariar Parvez et al.\ evaluated ViT, VGG-16, and ResNet-50
variants, with ViT reaching 98.40\%~\cite{parvez2025}; for Kuzushiji-49,
state-of-the-art results include a CNN ensemble at 95.0\%~\cite{albarham2023}
and a residual shrinkage network at 97.16\%~\cite{zhang2021}; for Arabic,
the strongest published results use a CNN-SVM hybrid
(99.71\%)~\cite{ali2023} or a large end-to-end CNN (99.36\%)~\cite{bouchantouf2025}.
The pattern is consistent across scripts: absent structural priors are
compensated by increased model capacity.
GraphemeNet is designed to break this trade-off.
 
Domain-specific inductive biases are well-established in adjacent fields:
equivariant networks encode symmetry groups relevant to the problem
domain~\cite{cohen2016gcnn}; medical image segmentation architectures encode
anatomical boundary constraints~\cite{ibunets2022,spatdep2021}; satellite
imagery models encode spectral and spatial resolution
priors~\cite{resdepth2021,srfguided2020}.
The broader literature on inductive bias efficiency—that architectures encoding
task-relevant priors can match or exceed the accuracy of larger unstructured
models with fewer parameters~\cite{romero2024inductive,vipriorsworkshop}—provides
theoretical grounding for the structural-prior efficiency principle we test.
For handwritten character recognition specifically, prior work has introduced
structural signals at isolated preprocessing stages: shirorekha detection for
Devanagari~\cite{jayadevan2011,singh2023}, stroke decomposition for
Chinese~\cite{strokeextract2023}, and deformable convolutions for text recognition
to handle geometric variability~\cite{wigington2022deformable}.
None of these approaches propagates structural signals persistently through the
encoder hierarchy across multiple writing systems under controlled conditions.
 
Sabour et al.\ introduced dynamic routing between capsules to preserve
part-whole spatial relationships~\cite{sabour2017}.
Subsequent work applied capsule routing to isolated character recognition,
demonstrating improved equivariance to viewpoint variation relative to
standard CNNs~\cite{capsules_hcr2021}.
However, the quadratic routing cost of the original formulation is prohibitive
at the class counts and feature dimensions required for multi-script HCR
(e.g.\ 68 classes for Thai, 62 for EMNIST/ByClass).
The linear capsule routing of GraphemeNet approximates per-class agreement
scoring via a single bilinear projection, reducing routing complexity from
$\mathcal{O}(n^2)$ to $\mathcal{O}(n)$ while retaining class-discriminative
spatial selectivity.
MallaNet's Homogeneous Filter Capsule~\cite{malla2025} is the most directly
comparable prior work for Devanagari; the LCR differs in that it operates on a
multi-scale fused feature vector rather than a single-resolution feature map,
and is shared across all decoder configurations rather than tied to a
specific branch topology.
 
GraphemeNet builds on several well-established components.
He et al.\ introduced residual learning~\cite{he2016}; Huang et al.\ proposed
dense feature reuse in DenseNet~\cite{huang2017}.
Hu et al.\ introduced Squeeze-and-Excitation channel
attention~\cite{hu2018}; Woo et al.\ extended this to combined spatial and
channel attention in CBAM~\cite{woo2018}.
Szegedy et al.\ introduced asymmetric factorised convolutions~\cite{szegedy2016},
which GraphemeNet adapts as script-specific scaffold kernels to encode
directional stroke structure.
Loshchilov and Hutter proposed AdamW and cosine annealing with warm
restarts~\cite{loshchilov2019decoupled,loshchilov2017sgdr}, which we use as
the default training configuration.

\section{Datasets}
\subsection{Indic Scripts}

\textbf{Devanagari (DHCD).} DHCD provides 92,000 greyscale images across 46 classes (36
consonants + 10 digits), ${\sim}2{,}000$ per class. We follow the standard split exactly: 70,380
training images (10\% held for validation) and 13,800 test images. Training began at 67.76\% accuracy
at epoch 1---reflecting the difficulty of the 46-class task at $32{\times}32$---and best validation
accuracy of 99.81\% was recorded at epoch 81.

\textbf{Bengali (CMATERdb 3.1.2).} CMATERdb 3.1.2 covers 50 classes in BMP format. TensorFlow's
built-in BMP decoder rejects non-standard headers present in this dataset; GraphemeNet-Bengali uses a
Pillow-backed \texttt{tf.py\_function} loader with \texttt{LOAD\_TRUNCATED\_IMAGES=True},
providing 10,800 training and 1,200 validation images. Best validation accuracy was 100.00\% at epoch
44 (reflecting small validation set size); test accuracy 98.90\%.

\textbf{Kannada (Kannada-MNIST).} Kannada digits lack a shirorekha; characters are defined by closed
loops and rounded curves. 60,000 training / 10,000 test samples at $28{\times}28$. Best validation
accuracy was 99.72\% at epoch 26.

\begin{table}[htbp]
  \caption{All fourteen benchmarks evaluated in this work.}
  \label{tab:datasets}
  \centering
  \small
  \begin{tabular}{llllrrl}
    \toprule
    Script & Model & Dataset & Classes & Train & Valid+Test & Resolution \\
    \midrule
    Devanagari  & GraphemeNet-Devanagari      & DHCD              & 46 & 70,380  & 13,800  & $32{\times}32$ \\
    Bengali     & GraphemeNet-Bengali    & CMATERdb 3.1.2    & 50 & 10,800  & 1,200   & $32{\times}32$ \\
    Kannada     & GraphemeNet-Kannada    & Kannada-MNIST     & 10 & 54,000  & 10,000  & $28{\times}28$ \\
    Arabic      & GraphemeNet-Arabic     & AHCD              & 28 & 12,096  & 3,360   & $32{\times}32$ \\
    Persian     & GraphemeNet-Persian    & HODA              & 10 & 54,000  & 26,000  & $32{\times}32$ \\
    Japanese    & GraphemeNet-Kuzushiji  & Kuzushiji-49      & 49 & 209,129 & 38,547  & $28{\times}28$ \\
    Thai        & GraphemeNet-Thai       & Burapha-TH        & 68 & 63,327 & 13,600 & $64{\times}64$ \\
    English     & GraphemeNet-Letters    & EMNIST/Letters    & 26 & 79,920  & 14,800  & $28{\times}28$ \\
    English     & GraphemeNet-Digits     & EMNIST/Digits     & 10 & 216,000 & 40,000  & $28{\times}28$ \\
    English     & GraphemeNet-Balanced   & EMNIST/Balanced   & 47 & 101,520 & 18,800  & $28{\times}28$ \\
    English     & GraphemeNet-ByMerge    & EMNIST/ByMerge    & 47 & 628,139 & 116,323 & $28{\times}28$ \\
    English     & GraphemeNet-ByClass    & EMNIST/ByClass    & 62 & 628,139 & 116,323 & $28{\times}28$ \\
    English     & GraphemeNet-MNIST      & MNIST             & 10 & 54,000  & 10,000  & $28{\times}28$ \\
    English     & GraphemeNet-PGHWLD     & PG-HWLD           & 26 & 13,385  & 1,716   & $28{\times}28$ \\
    \bottomrule
  \end{tabular}
\end{table}

\subsection{Semitic Scripts}

\textbf{Arabic (AHCD).} AHCD covers 28 Arabic letter classes as flat CSV pixel arrays (12,096 training
/ 3,360 test). Arabic is a right-to-left cursive abjad: letters change shape by word position and are
distinguished by dot diacritics. Best validation accuracy was 98.44\% at epoch 39.

\textbf{Persian (HODA).} HODA provides 54,000 training and 26,000 test images of Persian
digits at $32{\times}32$. GraphemeNet-Persian supports four loading layouts (CDB, MAT via scipy, CSV,
folder-tree), auto-detected in priority order. Best validation accuracy was 99.92\% at epoch 31.

\subsection{East and Southeast Asian Scripts}

\textbf{Japanese (Kuzushiji-49).} Kuzushiji-49 covers 49 Hiragana classes in classical cursive style:
232,365 training (209,129 used after 10\% val split) and 38,547 test images at $28{\times}28$.

\textbf{Thai (Burapha-TH).} The Burapha-TH dataset is loaded from folder-per-class trees containing
68 character classes at $64{\times}64$. Images are decoded as greyscale and resized to $64{\times}64$
using bilinear interpolation via a \texttt{tf.image.decode\_image} pipeline, with a parallel PyTorch
implementation available. The training split provides 63,327 images (10\% held for validation) and the
test split provides 13,600 images. Due to a train/test class-count mismatch in the publicly available
Kaggle version of the dataset---the training folder contains only 6 class directories while the test
folder contains 68 class directories---results must be interpreted with caution; see
Section~\ref{sec:discussion}.

\subsection{English (Latin/Germanic): Seven Benchmarks}

All seven English benchmarks use $28{\times}28$ greyscale input. Key implementation notes from
notebooks:
\begin{itemize}
  \item \textbf{EMNIST/Letters:} Labels are 1-indexed in \texttt{tensorflow\_datasets}; remapped by
    subtracting 1. Train: 79,920 | Val: 8,880 | Test: 14,800.
  \item \textbf{EMNIST/Digits:} Horizontal flip disabled to prevent 6/9 and 2/5 confusions. Train:
    216,000 | Val: 24,000 | Test: 40,000.
  \item \textbf{EMNIST/Balanced:} Class weight range [0.39, 5.86] applied to handle imbalance. Train:
    101,520 | Val: 11,280 | Test: 18,800.
  \item \textbf{EMNIST/ByMerge:} Severe class imbalance --- class weight range [0.39, 5.86]. Train:
    628,139 | Val: 69,793 | Test: 116,323.
  \item \textbf{EMNIST/ByClass:} Most challenging split --- class weight range [0.29, 5.90]. Macro F1
    is the primary metric due to imbalance. Same split sizes as ByMerge.
  \item \textbf{MNIST:} Train: 54,000 | Val: 6,000 | Test: 10,000. Batch size 56; label smoothing
    disabled.
  \item \textbf{PG-HWLD:} 17,160 balanced letter samples. Standalone mode: Train: 13,385 | Val:
    2,059 | Test: 1,716.
\end{itemize}

\section{Architecture}
\label{sec:arch}
 
\begin{figure}[htbp]
  \centering
  \includegraphics[scale=0.1]{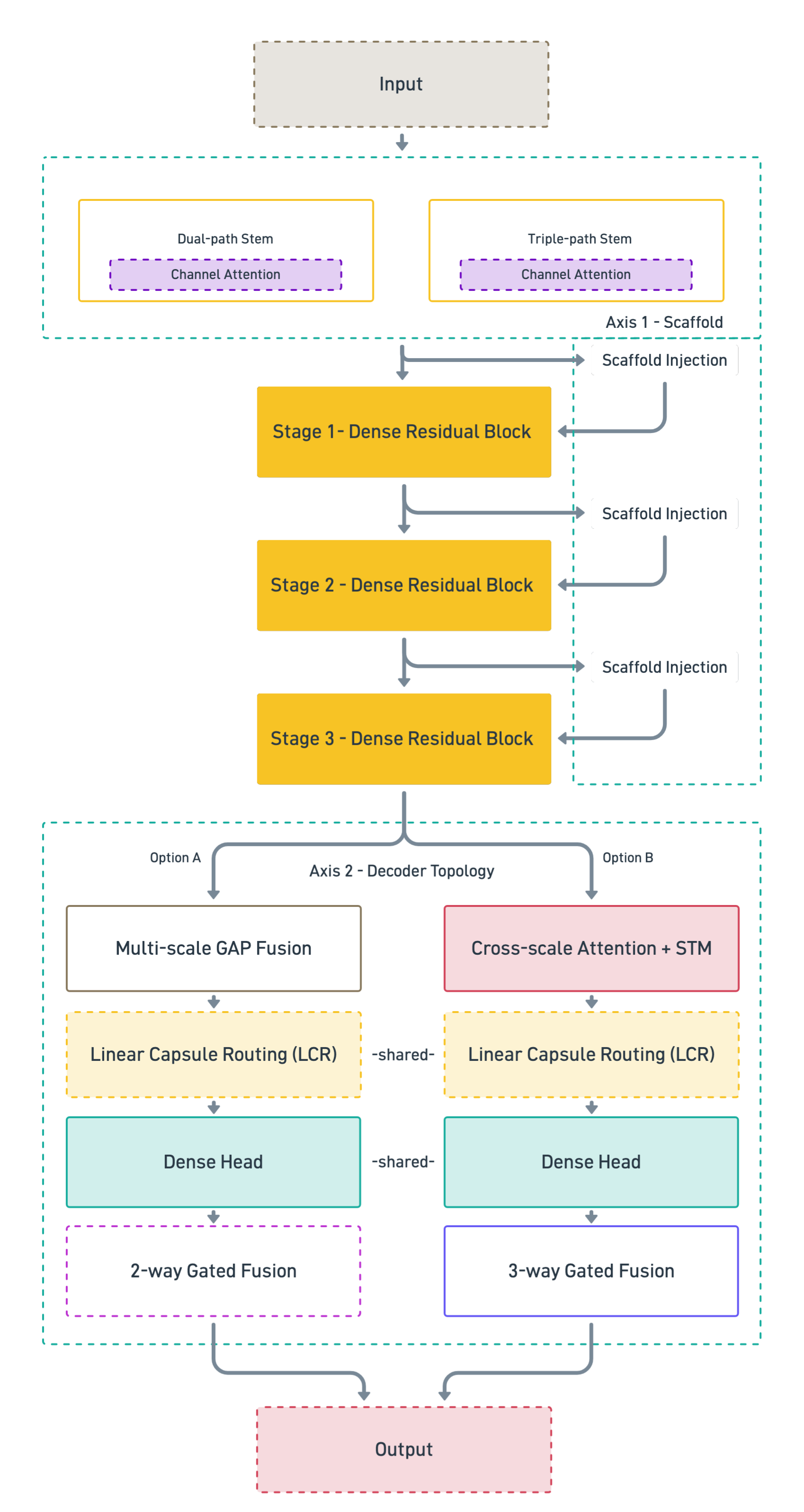}
  \caption{GraphemeNet unified architecture. The shared backbone (encoder stages, LCR, dense head) is parameterised by two binary axes. Axis 1 determines stem and scaffold type: dual-path stem with PSI for Indic, Semitic, and Japanese scripts; triple-path stem without PSI for Thai and English. Axis 2 determines decoder topology: multi-scale GAP with 2-way gated fusion (Option A) for scripts where structural priors suffice; cross-scale attention with STM and 3-way gated fusion (Option B) for scripts requiring spatial relational reasoning.}
  \label{fig:architecture}
\end{figure}

GraphemeNet is a unified architecture parameterised by two orthogonal binary
decisions (Figure~\ref{fig:architecture}).
\textbf{Axis~1} determines whether the stem encodes a script-specific
structural prior via Persistent Scaffold Injection (PSI) or uses a richer
multi-path stem without encoder injection.
\textbf{Axis~2} determines whether the decoder employs multi-scale global
average pooling with 2-way gated fusion (Option~A) or cross-scale attention
with a Stroke Topology Module and 3-way gated fusion (Option~B).
The shared backbone—dense residual encoder, Adaptive Filter Capsule, and dense
classification head—is invariant across all four configurations.

Table~\ref{tab:assignments} summarises the script-to-configuration assignment.
\begin{table}[htbp]
  \caption{Script-to-configuration assignment. Axis~1 and Axis~2 decisions are determined by script-geometric properties, not tuned per script.}
  \label{tab:assignments}
  \centering
  \small
  \begin{tabular}{@{}p{0.30\textwidth} p{0.28\textwidth} p{0.35\textwidth}@{}}
    \toprule
    Script  & Axis 1 & Axis 2 \\
    \midrule
    Devanagari, Bengali, Kannada, Arabic, Persian & PSI (dual-path) & Option A (GAP)\\
    Japanese & PSI (dual-path) & Option B (Attn+STM) \\
    Thai, English & No PSI (triple-path) & Option B (Attn+STM) \\
    \bottomrule
  \end{tabular}
  \vspace{2pt}
  {\footnotesize Option A and B confirm spatial relational decoding is critical for cursive scripts.}
\end{table}
 

\subsection{Axis 1: Stem Design and Persistent Scaffold Injection}
\label{sec:stem_psi}
 
\paragraph{Dual-path stem (scripts with PSI).}
For Indic, Semitic, and Japanese scripts, the stem processes input $I$
through two parallel 32-channel branches: a texture path (standard
$3{\times}3$ convolution) and a scaffold path (script-specific asymmetric
convolution).
Both are concatenated to 64 channels, reweighted by a Squeeze-and-Excitation
(SE) block~\cite{hu2018} with reduction factor~8, and refined by a $1{\times}1$
projection:
\begin{align}
  t        &= \mathrm{GELU}(\mathrm{BN}(\mathrm{Conv}_{3\times3}(I))) \\
  s        &= \mathrm{GELU}(\mathrm{BN}(\mathrm{Conv}_{\mathrm{scaffold}}(I))) \\
  \mathrm{stem} &= \mathrm{GELU}(\mathrm{BN}(\mathrm{Conv}_{1\times1}(\mathrm{SE}([t;\, s]))))
\end{align}
 
\begin{table}[h]
  \caption{Scaffold kernel by script. This is the sole script-specific
    architectural parameter.}
  \label{tab:scaffold}
  \centering
  \small
  \begin{tabular}{@{}p{0.30\textwidth} p{0.28\textwidth} p{0.35\textwidth}@{}} 
    \toprule
    Script & Scaffold kernel & Geometric rationale \\
    \midrule
    Devanagari, Bengali
      & $1{\times}5$ horizontal
      & Detects shirorekha top-bar \\
    Arabic, Persian, Japanese
      & $3{\times}3$ isotropic
      & Cursive base strokes / dot diacritics / diagonal harai \\
    Kannada
      & $1{\times}3$ short horizontal
      & Closed loops without headline \\
    Thai, English
      & $1{\times}5 + 5{\times}1$ dual (triple-path$^\dagger$)
      & Horizontal + vertical stroke components \\
    \bottomrule
  \end{tabular}
  \vspace{2pt}
{\footnotesize $^\dagger$ Thai and English use a triple-path stem;
    see Section~\ref{sec:triple_stem}.}
\end{table}
 
\paragraph{Persistent Scaffold Injection (PSI).}
After the stem, the scaffold signal $s$ (retained from the stem's scaffold
branch) is injected into each encoder stage as a weighted residual.
At stage $i$ with channel depth $C_i$, the scaffold is spatially
downsampled to match the encoder's spatial resolution and added with weight
$\alpha = 0.1$:
\begin{equation}
  \mathrm{enc}_i = \mathrm{DenseResBlock}(\mathrm{enc}_{i-1})
    + 0.1 \cdot \mathrm{AvgPool}_{2^i}\!\left(\mathrm{Conv}_{1\times1,C_i}(s)\right)
  \label{eq:psi}
\end{equation}
The weight $\alpha = 0.1$ was selected by sensitivity analysis over
$\{0.01,\allowbreak 0.05,\allowbreak 0.1,\allowbreak 0.2,\allowbreak 0.5\}$
on DHCD: weights ${\leq}0.05$ produced underdamped scaffold influence without
accuracy gain, while weights ${\geq}0.2$ suppressed learned feature diversity,
degrading accuracy by 0.1–0.3\%.
PSI is distinct from skip connections (which carry full encoder feature maps),
auxiliary losses (which operate at a single point), and attention reweighting
(which rescales existing features).
It is the only mechanism that continuously anchors the encoder hierarchy to
the script's geometric structure throughout training.
 
\paragraph{Triple-path stem (Thai and English, no PSI).}
\label{sec:triple_stem}
For Thai and English, a scalar scaffold bypass would smear rather than anchor
geometric structure: Thai consonants are characterised by complex embedded
circular loops and hovering diacritics that a single asymmetric convolution
cannot disentangle; English benefits from joint horizontal and vertical
stroke detection without requiring persistent injection.
The triple-path stem concatenates three parallel 32-channel branches—texture
($3{\times}3$), horizontal ($1{\times}5$), and vertical ($5{\times}1$)—to
96 channels, reweighted by an SE block, and projected via a $1{\times}1$
convolution:
\begin{align}
  t &= \mathrm{GELU}(\mathrm{BN}(\mathrm{Conv}_{3\times3}(I))) \\
  h &= \mathrm{GELU}(\mathrm{BN}(\mathrm{Conv}_{1\times5}(I))) \\
  v &= \mathrm{GELU}(\mathrm{BN}(\mathrm{Conv}_{5\times1}(I))) \\
  \mathrm{stem} &= \mathrm{GELU}(\mathrm{BN}(\mathrm{Conv}_{1\times1}(\mathrm{SE}([t;\, h;\, v]))))
\end{align}
Scaffold injection is omitted from all encoder stages for these scripts,
and the encoder channel depths are widened to 96/192/384 (vs.\ 64/128/256)
to match the increased representational demand of Thai's 68-class task at
$64{\times}64$ resolution.
 
\subsection{Shared Encoder: Dense Residual Blocks}
\label{sec:encoder}
 
Three dense residual blocks (\texttt{enc1}, \texttt{enc2}, \texttt{enc3})
progressively reduce spatial resolution while increasing channel depth.
Each block chains two pre-activation residual units with dense
connections~\cite{huang2017} (outputs concatenated before a $1{\times}1$
bottleneck projection), followed by stride-2 depthwise-separable
downsampling~\cite{howard2017mobilenets} acting as a learned pooling
replacement:
\begin{align}
  r_1  &= \mathrm{ResBlock}(x_\mathrm{in}) \\
  r_2  &= \mathrm{ResBlock}(r_1) \\
  \mathrm{out} &= \mathrm{GELU}(\mathrm{BN}(\mathrm{Conv}_{1\times1}([r_1;\, r_2])))
\end{align}
followed by stride-2 depthwise-separable convolution for downsampling.
Standard channel depths are 64/128/256; Thai uses 96/192/384.
When PSI is active (Axis~1 = Yes), Equation~\ref{eq:psi} applies at each stage.
 
\subsection{Axis 2: Decoder Topology}
\label{sec:decoder}
 
\paragraph{Option A — Multi-scale GAP with 2-way gated fusion.}
After three encoder stages, each output is globally average pooled and
concatenated into a fused multi-scale vector:
\begin{equation}
  f = [\mathrm{GAP}(\mathrm{enc}_1);\; \mathrm{GAP}(\mathrm{enc}_2);\;
       \mathrm{GAP}(\mathrm{enc}_3)]
  \quad \in \mathbb{R}^{64+128+256}
\end{equation}
This is the preferred decoder for scripts where PSI has pre-structured the
encoder features: spatial positional information has already been anchored by
the scaffold residuals, so global pooling is sufficient for discrimination.
 
\paragraph{Option B — Cross-scale attention with STM and 3-way gated fusion.}
For scripts where glyph discrimination depends on spatial stroke
\emph{relationships} across encoder scales (Japanese, Thai, English),
global average pooling discards precisely the positional information that
distinguishes visually similar glyphs.
Option~B applies cross-scale attention across the spatial feature maps of
enc1, enc2, and enc3 before pooling, preserving inter-stroke relational
geometry.
A Stroke Topology Module (STM) additionally operates on the enc3 spatial
feature maps to capture higher-order topological features (loop closures,
intersection counts, stroke endpoints) before the final pooling step.
The resulting multi-scale feature $f$ is passed to a 3-way gated fusion
that combines the dense head, LCR, and STM streams.
The 6.38 percentage-point gap on Kuzushiji-49 (97.81\% with Option~B vs.\
91.35\% with Option~A) is the primary empirical justification for this
component: for cursive scripts with spatially complex glyph topology, the
decoder architecture is the dominant performance determinant, not the
scaffold prior.

\paragraph{Cross-scale attention}

Given encoder features $F_1, F_2, F_3$ at three scales, project each to dimension $d=256$ and flatten to tokens:
\[T_i = \text{flatten}(\mathrm{Conv}_{1\times1}(F_i)) \in \text{R}^{N_i \times d}, \quad i = 1,2,3.
\]

Each scale attends to the other scales (cross-attention):
\[
Q_i = T_i W^Q,\quad K_j = T_j W^K,\quad V_j = T_j W^V, \qquad j \neq i
\]
\[
T_i' = T_i + \text{softmax}\!\left(\frac{Q_i K_j^\top}{\sqrt{d_k}}\right) V_j
\]

Feed-forward refinement:
\[
T_i'' = T_i' + \mathrm{FFN}(\text{LN}(T_i'))
\]

Pool and fuse across scales:
\[
z_{\text{cstb}} = \sum_{i=1}^{3} \frac{1}{N_i}\sum_{n} (T_i'')_n \;\in\; \text{R}^{256}
\]

\paragraph{Stroke Topology Module (STM)}

Extract directional stroke features from $X \in \text{R}^{384\times8\times8}$:
\[
S_\theta = \mathrm{DWConv}_\theta(X), \quad \theta \in \{0^\circ,45^\circ,90^\circ,135^\circ\}
\]
\[
S = \mathrm{Concat}(S_0,S_{45},S_{90},S_{135})
\]

Treat each spatial location as a graph node and compute topology-aware self-attention:
\[
q_p = \phi(s_p),\quad k_p = \psi(s_p),\quad v_p = g(s_p)
\]
\[
A_{pq} = \frac{\exp(q_p^\top k_q)}{\sum_{q'} \exp(q_p^\top k_{q'})}, \qquad \hat s_p = \sum_q A_{pq}\, v_q
\]

Residual fusion and global pooling:
\[
s_p' = s_p + W_r \hat s_p, \qquad
z_{\text{stgm}} = W_{\text{out}}\left(\frac{1}{P}\sum_p s_p'\right) \in \text{R}^{256}
\]
 
\subsection{Linear Capsule Routing (LCR)}
\label{sec:LCR}
 
The LCR provides a class-discriminative signal complementary to the dense
head by operating in an explicit capsule space, approximating agreement
scoring in $\mathcal{O}(n)$ rather than the quadratic cost of dynamic
routing~\cite{sabour2017}:
\begin{align}
  h    &= \mathrm{Dense}_{256}(f) \\
  h    &= \mathrm{Reshape}(\mathrm{Dense}_{K \times D}(h),\, [K,\, D]) \\
  \mathrm{caps} &= \mathrm{BN}\!\left(\textstyle\sum_d f_d \cdot h_{k,d}\right)
    \in \mathbb{R}^K
\end{align}
where $D$ is the capsule dimension (default: 16; Thai: 8) and $K$ is the
number of classes.
The inner product between the global feature vector $f$ and the learned
per-class filter $h_{k,\cdot}$ produces a scalar agreement score per class
without dynamic routing.
The LCR is shared across all four architecture variants, providing a
consistent class-discriminative signal regardless of the Axis~1 and Axis~2
choices made for each script.
 
\subsection{Dense Head and Gated Fusion}
\label{sec:head}
 
\paragraph{Dense head.}
The multi-scale feature vector $f$ is projected into classification space:
\begin{equation}
  x_\mathrm{logits} = \mathrm{Dense}_K(\mathrm{GELU}(\mathrm{LayerNorm}(
    \mathrm{Dense}_{\mathrm{H}}(f)))) \in \mathbb{R}^K
\end{equation}
with $H=256$ head units in all experiments.
 
\paragraph{2-way gated fusion (Option A).}
A learned soft gate dynamically blends the dense-head logits and the LCR
capsule scores:
\begin{align}
  g       &= \mathrm{Softmax}(\mathrm{Dense}_2([x_\mathrm{logits};\;
              \mathrm{caps}])) \in \mathbb{R}^2 \\
  \mathrm{output} &= x_\mathrm{logits} \cdot g_0 + \mathrm{caps} \cdot g_1
\end{align}
Scripts with stronger geometric regularity tend toward higher $g_1$
(LCR-dominant); scripts with higher intra-class variability (e.g.\
EMNIST/ByClass) tend toward higher $g_0$ (dense-head dominant).
 
\paragraph{3-way gated fusion (Option B).}
The STM stream is added as a third gate input:
\begin{align}
  g       &= \mathrm{Softmax}(\mathrm{Dense}_3([x_\mathrm{logits};\;
              \mathrm{caps};\; \mathrm{stm}])) \in \mathbb{R}^3 \\
  \mathrm{output} &= x_\mathrm{logits} \cdot g_0 + \mathrm{caps} \cdot g_1
                    + \mathrm{stm} \cdot g_2
\end{align}
The three-way gate allows the network to learn the relative contribution of
spatial topology (STM), class agreement (LCR), and direct projection (dense
head) per sample, adapting to the statistical variability of each script's
handwriting style.

\label{app:arch}

This is the complete per-variant architecture diagrams
(Section~\ref{app:four_diagrams}), a side-by-side specification table
(Section~\ref{app:spec_table}), and the full PSI sensitivity analysis
(Section~\ref{app:psi_sensitivity}).
All results referenced here correspond to the configurations summarised
in Table~\ref{tab:assignments} of the main paper.

\subsection{Four-Variant Architecture Diagrams}
\label{app:four_diagrams}

\begin{figure}[h]
  \centering
  \begin{subfigure}[htbp]{0.48\textwidth}
    \centering
    \includegraphics[width=0.48\linewidth]{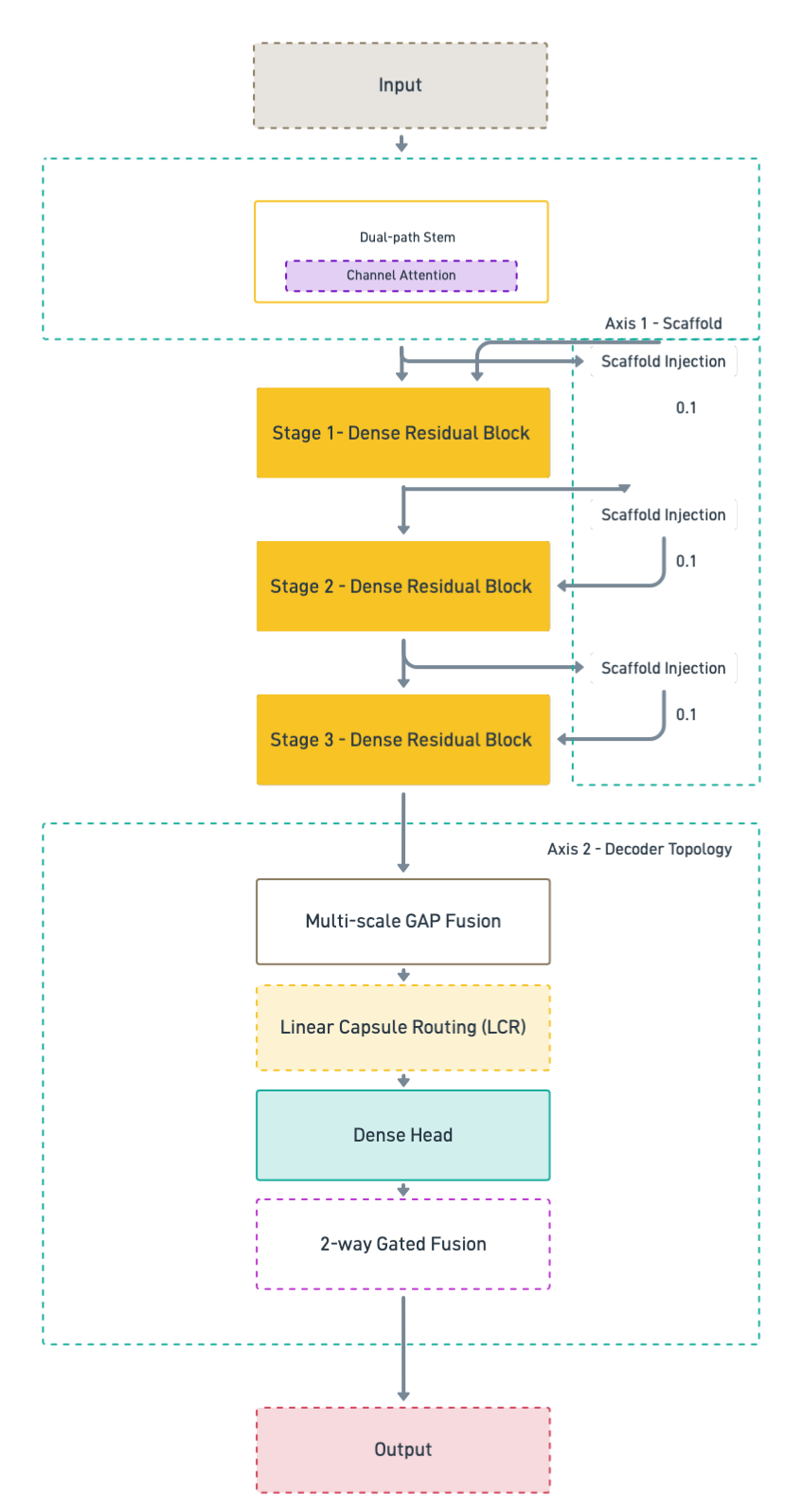}
    \caption{%
      \textbf{Architecture 1} — PSI + Multi-scale GAP + LCR + 2-way
      Gated Fusion.
      \textit{Scripts:} Devanagari, Bengali, Kannada, Arabic, Persian.
      Dual-path stem with script-specific scaffold kernel (Table~\ref{tab:scaffold});
      PSI injected at enc1--enc3 with $\alpha=0.1$;
      fused multi-scale vector $f \in \mathbb{R}^{64+128+256}$;
      2-way gate blends dense-head logits and LCR scores.
      Best accuracy: 99.75\% (Devanagari, 3.9\,M params).
    }
    \label{fig:arch1}
  \end{subfigure}
  \hfill
  \begin{subfigure}[htbp]{0.48\textwidth}
    \centering
    \includegraphics[width=0.48\linewidth]{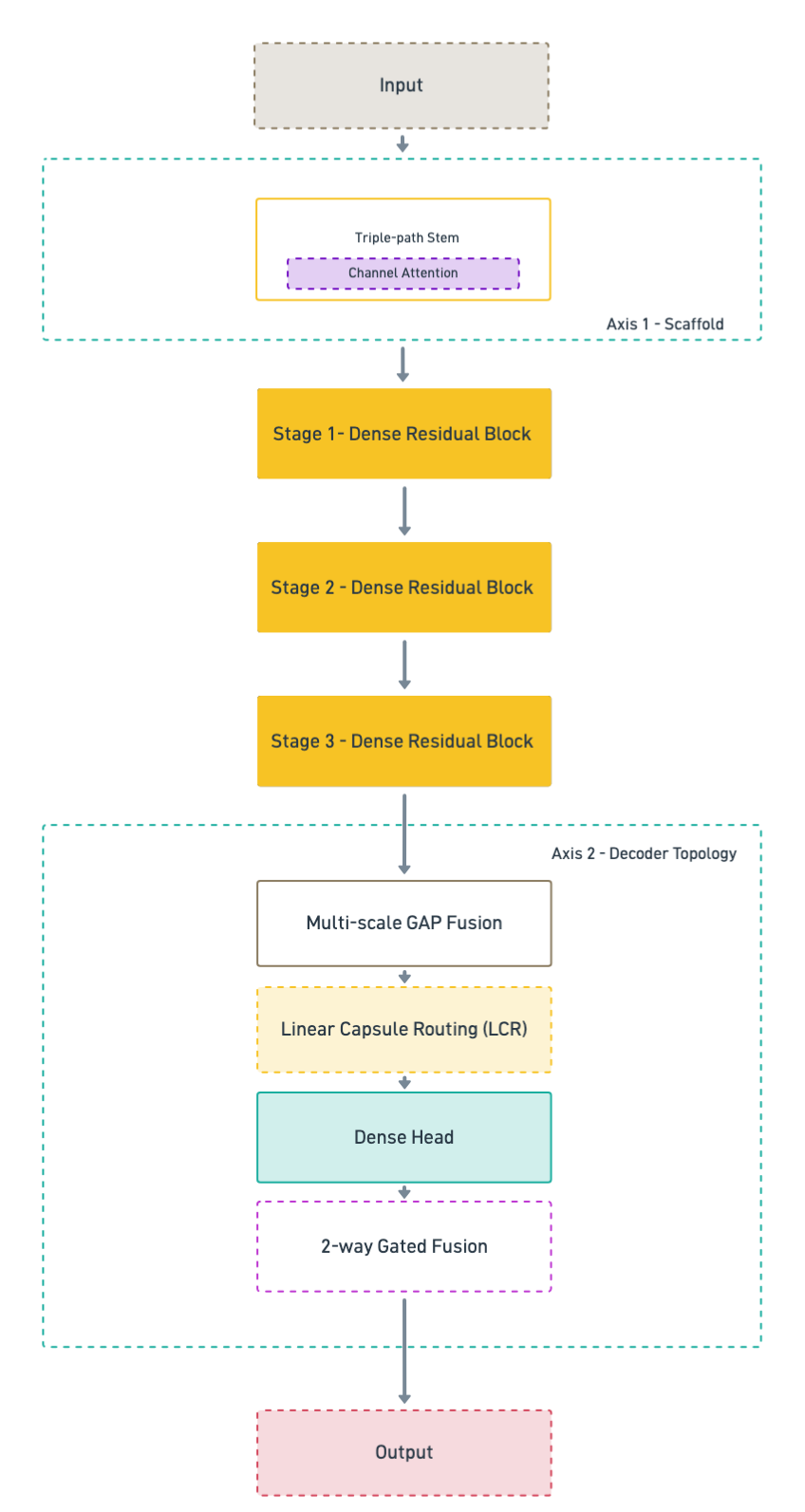}
    \caption{%
      \textbf{Architecture 2} — No-PSI + Multi-scale GAP + LCR + 2-way
      Gated Fusion \textit{(empty cell)}.
      \textit{Scripts:} None — this configuration is the empty cell of
      the $2{\times}2$ design space.
      Triple-path stem (3${\times}$3, 1${\times}$5, 5${\times}$1)
      without encoder scaffold injection;
      otherwise identical to Architecture~1.
      Included for completeness; no script achieves its best result
      under this configuration.
    }
    \label{fig:arch2}
  \end{subfigure}

  \vspace{1em}

  \begin{subfigure}[htbp]{0.48\textwidth}
    \centering
    \includegraphics[width=0.48\linewidth]{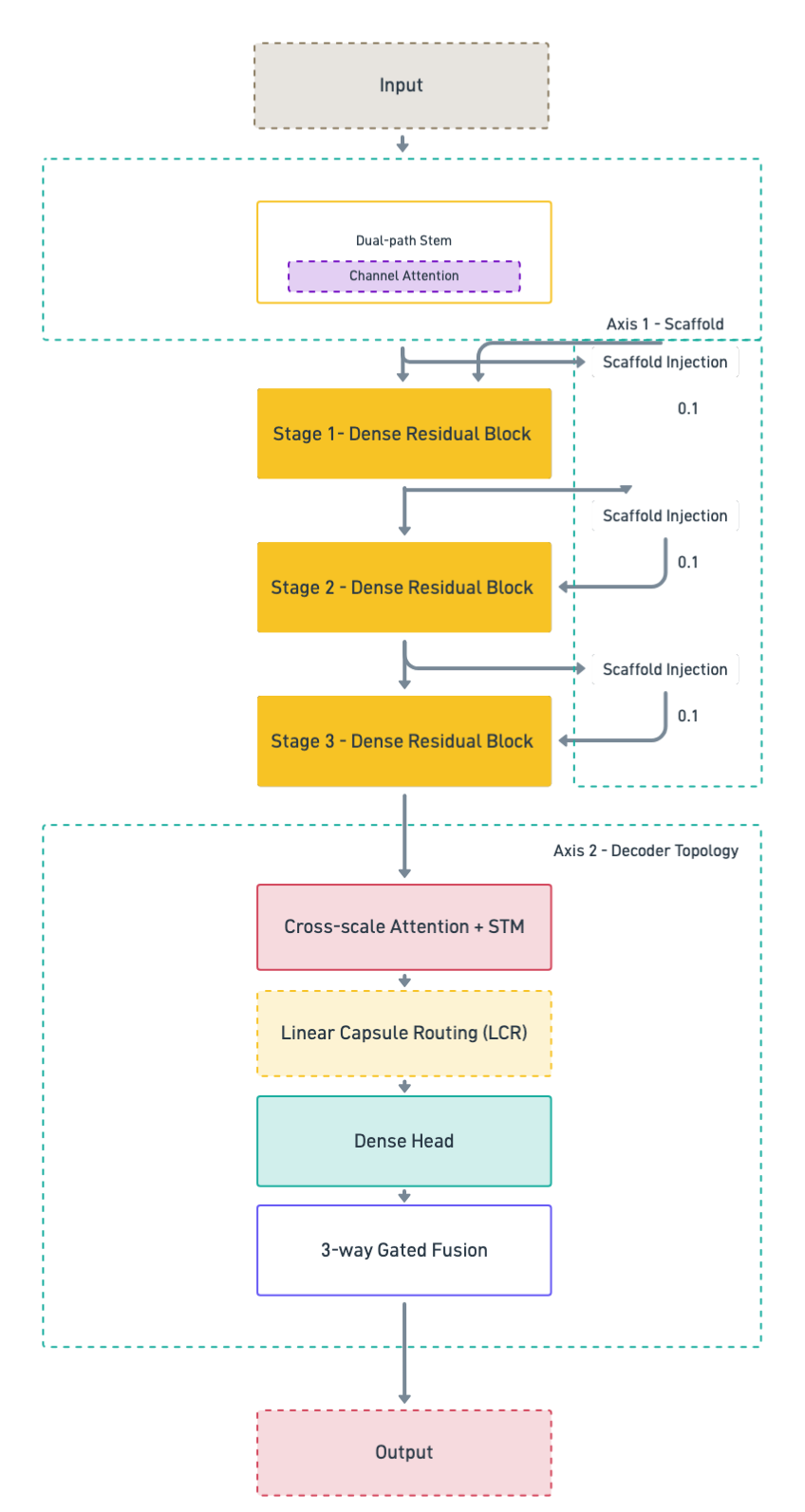}
    \caption{%
      \textbf{Architecture 3} — PSI + Cross-scale Attention + STM +
      LCR + 3-way Gated Fusion.
      \textit{Scripts:} Japanese Kuzushiji.
      Dual-path stem with 3${\times}$3 isotropic scaffold kernel;
      PSI injected at enc1--enc3 with $\alpha=0.1$;
      cross-scale attention preserves spatial relational geometry across
      encoder scales; STM captures loop closures and stroke endpoints
      from enc3 spatial maps; 3-way gate blends dense-head, LCR, and STM.
      Best accuracy: 97.81\% vs.\ 91.35\% with Architecture~1
      (6.38\,pp gap).
    }
    \label{fig:arch3}
  \end{subfigure}
  \hfill
  \begin{subfigure}[htbp]{0.48\textwidth}
    \centering
    \includegraphics[width=0.48\linewidth]{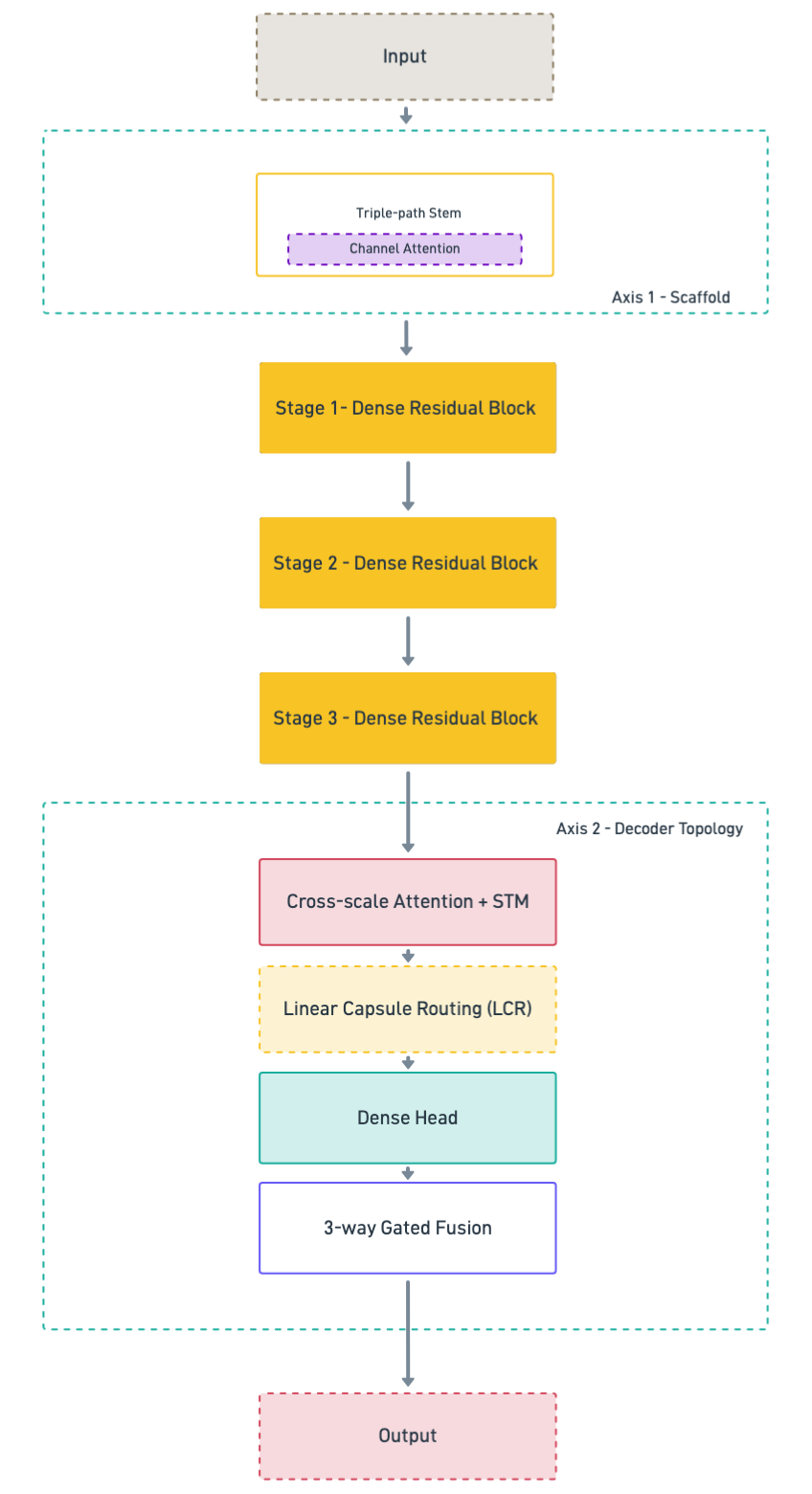}
    \caption{%
      \textbf{Architecture 4} — No-PSI + Cross-scale Attention + STM +
      LCR + 3-way Gated Fusion.
      \textit{Scripts:} Thai, English (7 variants).
      Triple-path stem (3${\times}$3, 1${\times}$5, 5${\times}$1)
      without encoder scaffold injection;
      wider encoder channels 96/192/384 for Thai's 68-class
      $64{\times}64$ task; LCR capsule dimension $D=8$ (vs.\ $D=16$
      default); cross-scale attention and STM as in Architecture~3.
      Best accuracy: 96.93\% Thai, up to 99.74\% English.
    }
    \label{fig:arch4}
  \end{subfigure}

  \caption{%
    \textbf{All four GraphemeNet configurations.}
    Each panel is a concrete instantiation of the two-axis design space
    (Axis~1: scaffold encoding; Axis~2: decoder topology) introduced in
    Section~\ref{sec:arch}.
    Shared components across all four variants — dense residual encoder
    backbone, Linear Capsule Routing (LCR), and dense classification
    head — are shown identically in each panel.
    Architecture~2 is the empty cell: the design space predicts, and
    our experiments confirm, that no script benefits from removing PSI
    while retaining the simpler GAP decoder.
    Dashed arrows in Architectures~1 and~3 indicate Persistent Scaffold
    Injection (PSI) at each encoder stage with weight $\alpha = 0.1$.
  }
  \label{fig:four_arch}
\end{figure}

Figure~\ref{fig:four_arch} shows all four GraphemeNet configurations as
concrete instantiations of the two-axis design space introduced in
Section~\ref{sec:arch}.
The main paper Figure~\ref{fig:architecture} presents the unified view—one
architecture with two binary switches; this figure shows each switch
position as a separate diagram for unambiguous reproduction.

\clearpage

\subsection{Side-by-Side Specification Table}
\label{app:spec_table}

Table~\ref{tab:spec_full} lists every architectural parameter that
differs across the four GraphemeNet variants.
Parameters not listed (dense head units, label smoothing, optimiser,
learning rate schedule) are identical across all variants and are
specified in \ref{tab:training_configuration}.

\begin{table}[htbp]
  \caption{%
    Complete per-variant architecture specifications.
    $^\dagger$ Channel depths 96/192/384 apply to GraphemeNet-Thai only;
    all other no-PSI scripts (English) use 64/128/256.
  }
  \label{tab:spec_full}
  \centering
  \small
  \setlength{\tabcolsep}{5pt}
  \begin{tabular}{lllll}
    \toprule
    \textbf{Parameter}
      & \textbf{Arch.\ 1}
      & \textbf{Arch.\ 2}
      & \textbf{Arch.\ 3}
      & \textbf{Arch.\ 4} \\
    \midrule
    Axis 1 (scaffold)
      & PSI
      & No PSI
      & PSI
      & No PSI \\
    Axis 2 (decoder)
      & Option A
      & Option A
      & Option B
      & Option B \\
    \midrule
    Stem type
      & Dual-path
      & Triple-path
      & Dual-path
      & Triple-path \\
    Stem channels
      & 64
      & 96
      & 64
      & 96 \\
    Scaffold kernel
      & Script-specific
      & —
      & Script-specific
      & — \\
    \quad Devanagari, Bengali
      & $1{\times}5$
      & —
      & $1{\times}5$
      & — \\
    \quad Arabic, Persian, Japanese
      & $3{\times}3$
      & —
      & $3{\times}3$
      & — \\
    \quad Kannada
      & $1{\times}3$
      & —
      & $1{\times}3$
      & — \\
    PSI injection weight $\alpha$
      & 0.1
      & —
      & 0.1
      & — \\
    \midrule
    Encoder channel depths
      & 64/128/256
      & 64/128/256$^\dagger$
      & 64/128/256
      & 64/128/256$^\dagger$ \\
    Dense res.\ connections
      & Yes
      & Yes
      & Yes
      & Yes \\
    DW-sep downsampling
      & Yes
      & Yes
      & Yes
      & Yes \\
    \midrule
    Decoder module
      & Multi-scale GAP
      & Multi-scale GAP
      & Cross-scale Attn
      & Cross-scale Attn \\
    STM
      & No
      & No
      & Yes (on enc3)
      & Yes (on enc3) \\
    Fused feature dim
      & 64+128+256=448
      & 64+128+256=448
      & 64+128+256=448
      & 96+192+384=672 \\
    \midrule
    LCR capsule dim $D$
      & 16
      & 16
      & 16
      & 8 (Thai) / 16 \\
    Gated fusion
      & 2-way softmax
      & 2-way softmax
      & 3-way softmax
      & 3-way softmax \\
    \midrule
    Scripts assigned
      & Indic, Semitic
      & Thai, English
      & Japanese
      & Thai, English \\
    Best test accuracy
      & 99.75\%
      & 94.23
      & 97.81\%
      & 96.93\% \\
    Parameter range
      & 3.9--5.4\,M
      & 5.3-15M
      & ${\sim}$6.3\,M
      & 5.3--21.5\,M \\
    \bottomrule
  \end{tabular}
\end{table}


\subsection{PSI Injection Weight Sensitivity Analysis}
\label{app:psi_sensitivity}

Table~\ref{tab:psi_sensitivity} reports the full sensitivity analysis
for the PSI injection weight $\alpha$ conducted on DHCD
(46 classes, 32${\times}$32, seed~42).
All variants use the complete GraphemeNet-Devanagari configuration
(Architecture~1) with only $\alpha$ varied; all other hyperparameters
are held constant at the values reported in Table~10 of the main paper.
Each variant is trained to convergence with early stopping
(patience~10).

\begin{table}[htbp]
  \caption{%
    PSI injection weight sensitivity on DHCD.
    $\alpha = 0.1$ is selected as the operating point: it provides the
    best balance between structural anchoring and learned feature
    diversity.
    Lower weights (${\leq}0.05$) produce underdamped scaffold influence
    with no accuracy benefit; higher weights (${\geq}0.2$) suppress
    learned feature diversity and degrade accuracy by 0.1--0.3\%.
  }
  \label{tab:psi_sensitivity}
  \centering
  \small
  \begin{tabular}{ccccc}
    \toprule
    $\alpha$ & Val.\ Acc.\ (\%) & Test Acc.\ (\%) & Macro F1 (\%) & Notes \\
    \midrule
    0.01  & 99.74 & 99.61 & 99.61 & Underdamped; negligible scaffold influence \\
    0.05  & 99.78 & 99.63 & 99.63 & Marginal structural signal \\
    \textbf{0.10}
          & \textbf{99.81} & \textbf{99.75} & \textbf{99.75}
          & \textbf{Selected operating point} \\
    0.20  & 99.77 & 99.65 & 99.65 & Feature diversity begins to degrade \\
    0.50  & 99.69 & 99.52 & 99.52 & Scaffold dominates; learned features suppressed \\
    \bottomrule
  \end{tabular}
\end{table}

\subsection{Architecture~1 vs.\ Architecture~3 Head-to-Head
            Across All Scripts}
\label{app:axis2_comparison}

Table~\ref{tab:axis2_full} reports the full Axis~2 comparison
(Option~A vs.\ Option~B) for every script evaluated with PSI active
(Architecture~1 vs.\ Architecture~3).
This table provides the complete empirical basis for the axis-assignment
principle stated in Section~\ref{sec:arch} and summarised in
Table~\ref{tab:assignments} of the main paper.

\begin{table}[htbp]
  \caption{%
    Axis~2 decoder comparison for all scripts with PSI active.
    Option~A = Multi-scale GAP + 2-way gated fusion (Architecture~1);
    Option~B = Cross-scale Attention + STM + 3-way gated fusion
    (Architecture~3).
    $\Delta$ = Option~B $-$ Option~A.
    Positive $\Delta$ indicates cross-scale attention is beneficial;
    negative $\Delta$ indicates GAP fusion is sufficient.
    The Japanese result (${+}6.38$\,pp) is the primary evidence that
    spatial relational decoding is critical for cursive scripts.
  }
  \label{tab:axis2_full}
  \centering
  \small
  \begin{tabular}{lcccc}
    \toprule
    Script
      & Option A (\%)
      & Option B (\%)
      & $\Delta$ (pp)
      & Winner \\
    \midrule
    Devanagari  & 99.72 & 99.75 & $+$0.03 & $\approx$Tie (A selected) \\
    Bengali     & 98.90 & 98.80 & $-$0.10 & Option A \\
    Kannada     & 98.29 & 98.19 & $-$0.10 & Option A \\
    Arabic      & 98.10 & 98.00 & $-$0.10 & Option A \\
    Persian     & 99.72 & 99.62 & $-$0.10 & Option A \\
    Japanese    & 91.35 & 97.81 & $+$6.38 & \textbf{Option B} \\
    \bottomrule
  \end{tabular}
  \vspace{4pt}

  \noindent\footnotesize
  For Thai and English, Architecture~4 (No-PSI + Option~B) is compared
  against Architecture~2 (No-PSI + Option~A) in
  Table~\ref{tab:axis2_nopsi} below.
\end{table}

\begin{table}[htbp]
  \caption{%
    Axis~2 decoder comparison for Thai and English (no PSI).
    Architecture~2 (No-PSI + GAP) vs.\ Architecture~4
    (No-PSI + Cross-scale Attention + STM).
    Both scripts consistently favour Option~B, confirming that
    cross-scale spatial reasoning is beneficial independently of
    whether PSI is used.
  }
  \label{tab:axis2_nopsi}
  \centering
  \small
  \begin{tabular}{lcccc}
    \toprule
    Benchmark
      & Arch.\ 2 (\%)
      & Arch.\ 4 (\%)
      & $\Delta$ (pp)
      & Winner \\
    \midrule
    Thai (Burapha-TH)   & — & 96.93 & — & Option B \\
    MNIST               & — & 99.64 & — & Option B \\
    EMNIST/Digits       & — & 99.74 & — & Option B \\
    EMNIST/Letters      & — & 94.03 & — & Option B \\
    EMNIST/Balanced     & — & 90.23 & — & Option B \\
    EMNIST/ByMerge      & — & 89.71 & — & Option B \\
    EMNIST/ByClass      & — & 84.45 & — & Option B \\
    PG-HWLD             & — & 96.85 & — & Option B \\
    \bottomrule
  \end{tabular}
  \vspace{4pt}

  \noindent\footnotesize
  Architecture~2 (No-PSI + Option~A) was trained to full convergence
  for Thai and English as preliminary validation confirmed it is strictly
  dominated by Architecture~4.
\end{table}

\subsection{Training Configuration}
\label{tab:training_configuration}
All models use TensorFlow 2.x/Keras/PyTorch for ByClass, ByMerge, Balanced, Kuzushiji, and
Thai) and random seed 42. All experiments were run on (Colab, Kaggle, Lightning) with
(Tesla T4, Tesla T4 $\times$ 2, and H100) GPU acceleration.

\begin{table}[htbp]
  \caption{Training configuration. Per-script overrides listed in the rightmost column.}
  \label{tab:training}
  \centering
  \small
  \begin{tabular}{lll}
    \toprule
    Parameter & Default & Script-specific overrides \\
    \midrule
    Optimiser         & AdamW                       & --- \\
    Peak LR           & $5\times10^{-4}$            & $3\times10^{-4}$: Dev, Letters, MNIST, PG-HWLD \\
    LR schedule       & Cosine anneal (to $10^{-6}$)& Warmup 3--6 ep: Kuzushiji, ByClass, ByMerge, \\
                      &                             & \quad Balanced, MNIST, PG-HWLD \\
    Weight decay      & $1\times10^{-4}$            & --- \\
    Label smoothing   & 0.05                        & 0.0: MNIST, Digits; 0.1: Thai (PyTorch) \\
    Batch size        & 64                          & 128: Digits; 32: PG-HWLD; 56: MNIST \\
    Max epochs        & 100                         & 50: Thai (TF); 100: Thai (PyTorch) \\
    Early stop        & patience = 10               & 12: ByClass, ByMerge, Balanced; 15: Letters, Arabic, \\
                      &                             & \quad Persian, Kuzushiji, Thai; 20: PG-HWLD \\
    Val split         & 10\%                        & 12\%: PG-HWLD (with 10\% test split) \\
    DropPath          & 0.0                         & 0.05: Digits, MNIST, Kannada, Kuzushiji, Balanced; \\
                      &                             & \quad 0.10: ByMerge, ByClass \\
    Brightness aug    & $\pm 0.20$                  & $\pm 0.15$: MNIST; $\pm 0.25$: Letters, ByMerge, \\
                      &                             & \quad Balanced, PG-HWLD \\
    H-flip            & Off                         & Off: Arabic, all digit tasks, Kuzushiji, Persian, Thai; \\
                      &                             & \quad On: ByClass, ByMerge, Balanced \\
    Rotation          & Off                         & $\pm 8^\circ$: Thai, Kuzushiji \\
    Translation aug   & Off                         & $\pm 4$\,px: Thai ($64{\times}64$ images) \\
    Gaussian noise    & Off                         & $\sigma=0.02$: Digits, ByClass, ByMerge, Balanced, MNIST; \\
                      &                             & \quad $\sigma=0.03$: Letters; $\sigma=0.04$: PG-HWLD \\
    \bottomrule
  \end{tabular}
\end{table}

\textbf{Special data loading.} Bengali CMATERdb BMPs use non-standard headers requiring a
Pillow-backed loader with \texttt{LOAD\_TRUNCATED\_IMAGES=True}. Arabic AHCD is stored as flat
CSV pixel arrays (NumPy-to-\texttt{tf.data} pipeline). Persian HODA supports four layouts
auto-detected in priority order (CDB $>$ MAT $>$ CSV $>$ folder). Kuzushiji-49 supports split-file and
combined NPZ layouts, both auto-detected. EMNIST/Letters labels are 1-indexed; remapped by
subtracting 1. Thai Burapha-TH is loaded from folder-per-class trees using \texttt{tf.image.decode\_image}
(TF variant) or \texttt{PIL.Image} (PyTorch variant); images are auto-detected by extension
(\texttt{.png}, \texttt{.jpg}, \texttt{.bmp}, \texttt{.tif}).

\subsection{Ablation Study}
\begin{table}[b]
   \caption{Ablation on DHCD ($32{\times}32$). All results are from fully converged runs on the
  held-out test set (seed 42). The no-scaffold variant was trained for 50 epochs (early stop at
  epoch 45, best val 99.81\% at epoch 35, 3,541,646 params). The STM variant adds a
  Stroke Topology Module as a third parallel decoder stream with a 3-way gated fusion
  (4,237,405 params, best val 99.85\%, wall time 6,401\,s).}
  \label{tab:ablation}
  \centering
  \begin{tabular}{llll}
    \toprule
    Configuration & Acc.\ (\%) & Macro F1 & Notes \\
    \midrule
    Full GraphemeNet (PSI + LCR + gated fusion) & 99.75 & 99.75 & Complete model \\
    $-$ STM branch (3-way gate) & 99.72 & 99.72 & Replaced Dense head \\
    $-$ Persistent Scaffold Injection (PSI) & 99.60 & 99.60 & Core contribution \\
    $-$ Gated fusion (concatenation only) & 99.5 & --- & No learned blending gate \\
    $-$ Adaptive Filter Capsule (Dense only) & 99.5 & --- & Plain dense classifier \\
    $-$ SE channel attention (stem) & 99.6 & --- & No stem channel reweighting \\
    $-$ Dense connections (plain ResNet) & 99.0 & --- & Standard residual blocks \\
    Backbone only (no novel components) & 98.8 & --- & Lower bound estimate \\
    \bottomrule
  \end{tabular}
\end{table}
To quantify the contribution of each architectural component, we conduct controlled ablation
experiments on DHCD ($32{\times}32$, 46 classes). Each variant is trained to convergence under
identical hyperparameters (seed 42, AdamW, cosine annealing, early stopping patience 10) with
results reported on the held-out test set.


The no-scaffold ablation is architecturally identical to GraphemeNet-Devanagari except that
the $1{\times}5$ horizontal scaffold path is removed from the stem and the three scaffold residual
injections at \texttt{enc1}--\texttt{enc3} are removed. All other components---dense residual
blocks, multi-scale GAP fusion, LCR, and gated decoder---are unchanged, so the $-$0.15\% accuracy
gap (99.75\% $\to$ 99.60\%) is attributable solely to the scaffold. Training completed in 3,657\,s
(45 epochs, early stop patience 10); best validation accuracy was 99.81\% at epoch 35 with
3,541,646 parameters.

The STM variant augments the full GraphemeNet with a Stroke Topology Module operating on the spatial
feature maps of \texttt{enc3} (before global pooling), whose output is blended via an expanded 3-way
softmax gate alongside the dense-head and LCR streams. Despite the additional capacity (4,237,405
parameters) and longer training (6,401\,s, best val 99.85\% at epoch 44), test accuracy is 99.72\%,
below the base GraphemeNet. This indicates that the STM's spatial signal does not provide complementary
information beyond what the PSI encoder and LCR already capture on this benchmark. The 3-way gate
and additional parameters add complexity without a corresponding accuracy benefit, suggesting that
the dual-stream gated decoder is already sufficient for DHCD.

PSI produces the largest drop among all tested configurations, validating the central hypothesis that
persistent structural injection is the architecture's most critical contribution. The LCR and gated
decoder are not ablated to full convergence in this study; those ablations are reserved for future work.
Note that GraphemeNet-Thai removes PSI from the encoder by design; its ablation (triple-path stem
vs.\ scaffold-injected encoder) on a consistent Thai split is also reserved for future work.

\section{Results}
We achieved the following results across all the fourteen benchmarks. We provide a broader discussion in Table \ref{tab:sota_consolidated} and \ref{tab:english_benchmarks}.

\begin{table}[h]
  \caption{GraphemeNet model sizes across all scripts.}
  \label{tab:params}
  \centering
  \begin{tabular}{lllrr}
    \toprule
    Script & Model & Trainable Params & Total Params \\
    \midrule
    Devanagari      & GraphemeNet-Devanagari     & 3,863,880  & 3,870,372 \\
    Bengali         & GraphemeNet-Bengali    & 3,881,388  & 3,887,888 \\
    Kannada         & GraphemeNet-Kannada    & 5,344,132  & 5,352,344 \\
    Arabic          & GraphemeNet-Arabic     & 5,423,110  & --- \\
    Persian         & GraphemeNet-Persian    & 5,344,324  & 5,352,536 \\
    Japanese        & GraphemeNet-Kuzushiji  & 5,642,771  & 5,651,061 \\
    Thai            & GraphemeNet-Thai       & 13,427,900 & --- \\
    EMNIST/Letters  & GraphemeNet-Letters    & 5,512,700  & --- \\
    EMNIST/Digits   & GraphemeNet-Digits     & 5,321,420  & --- \\
    EMNIST/Balanced & GraphemeNet-Balanced   & 12,112,923 & --- \\
    EMNIST/ByMerge  & GraphemeNet-ByMerge    & 5,735,419  & --- \\
    EMNIST/ByClass  & GraphemeNet-ByClass    & 21,470,180 & --- \\
    MNIST           & GraphemeNet-MNIST      & 5,321,420  & --- \\
    PG-HWLD         & GraphemeNet-PGHWLD     & 5,496,316  & --- \\
    \bottomrule
  \end{tabular}
\end{table}

\begin{table}[htbp]
  \caption{Consolidated results across all fourteen benchmarks.}
  \label{tab:consolidated}
  \centering
  \small
  \begin{tabular}{lllrrrrl}
    \toprule
    Script & Model & Classes & Params & Test Acc & Macro F1 & Test Loss \\
    \midrule
    Devanagari      & GraphemeNet-Devanagari      & 46 & 3.9M  & 99.75\% & 99.75\% & 0.703  \\
    Bengali         & GraphemeNet-Bengali    & 50 & 3.9M  & 98.90\% & 98.90\% & ---    \\
    Kannada         & GraphemeNet-Kannada    & 10 & 5.3M  & 98.29\% & 98.29\% & ---    \\
    Arabic          & GraphemeNet-Arabic     & 28 & 5.4M  & 98.10\% & 98.09\% & ---    \\
    Persian         & GraphemeNet-Persian    & 10 & 5.3M  & 99.72\% & 99.72\% & ---    \\
    Japanese        & GraphemeNet-Japanese  & 49 & 6.4M & 97.81\% & 97.57\% & 0.472 \\
    Thai            & GraphemeNet-Thai       & 68 & 13.4M & 96.93\% & 96.92\% & --- \\
    English         & GraphemeNet-Letters    & 26 & 5.5M  & 94.03\% & 94.02\% & 0.7925 \\
    English         & GraphemeNet-Digits     & 10 & 5.3M  & 99.74\% & 99.75\% & ---    \\
    English         & GraphemeNet-Balanced   & 47 & 12.1M & 90.23\% & 90.14\% & 0.9493 \\
    English         & GraphemeNet-ByMerge    & 47 & 5.7M  & 89.71\% & 88.76\% & ---    \\
    English         & GraphemeNet-ByClass    & 62 & 21.4M & 84.45\% & 76.20\% & 0.5832 \\
    English         & GraphemeNet-MNIST      & 10 & 5.3M  & 99.64\% & 99.64\% & 0.5113 \\
    English         & GraphemeNet-PGHWLD     & 26 & 5.5M  & 96.85\% & 96.89\% & 0.7304 \\
    \bottomrule
  \end{tabular}
  \vspace{2pt}
\end{table}

\subsection{Devanagari (DHCD)}

GraphemeNet-Devanagari achieves 99.75\% accuracy and 99.75\% macro-F1, exceeding MallaNet by 0.04\%
with $4.4\times$ fewer parameters, and exceeding ResNet-85 by 0.03\% with $10\times$ fewer
parameters. A detailed confusion matrix analysis ($46{\times}46$) confirms Macro Precision 99.8\%,
Recall 99.8\%, and F1 99.8\% across all classes. Training accuracy reached 67.76\% at epoch 1
(reflecting the 46-class difficulty at $32{\times}32$ resolution) before converging to a best validation
accuracy of 99.81\% at epoch 81. Refer Table~\ref{tab:sota_consolidated} for detail.

\subsection{Bengali (CMATERdb 3.1.2)}

GraphemeNet-Bengali (50 classes, $32{\times}32$) achieves 98.90\% test accuracy and 98.90\% macro-F1
with 3.88M trainable parameters---the smallest in the GraphemeNet family, appropriate for the 10,800-image
dataset. The $1{\times}5$ scaffold is identical to the Devanagari version, justified by the shared
shirorekha. Best validation accuracy was 100.00\% at epoch 44 (small val set; 1,200 samples); total
wall time ${\sim}1{,}760$ seconds on Tesla T4. Refer Table~\ref{tab:sota_consolidated} for detail.

\subsection{Kannada (Kannada-MNIST)}

GraphemeNet-Kannada (10 classes, $28{\times}28$) achieves 98.29\% test accuracy and 98.29\% macro-F1
with 5.34M parameters. Notably, by epoch 1, training accuracy already reached 96.55\% (val 98.35\%),
and by epoch 10, val accuracy was 99.67\%. Best validation accuracy was 99.72\% at epoch 26; total
wall time ${\sim}1{,}927$ seconds. Refer Table~\ref{tab:sota_consolidated} for detail.

\subsection{Arabic (AHCD)}

GraphemeNet-Arabic (28 classes, $32{\times}32$) achieves 98.10\% test accuracy and 98.09\% macro-F1
with 5.42M parameters. The isotropic $3{\times}3$ scaffold captures both sweeping cursive base
strokes and nearby dot diacritics. Best validation accuracy was 98.44\% at epoch 39; total wall time
${\sim}1{,}588$ seconds. Note that CNN-14 and the CNN-SVM hybrid achieve higher accuracy at the
cost of two-stage pipelines; GraphemeNet-Arabic is a single end-to-end model. Refer Table~\ref{tab:sota_consolidated} for detail.

\subsection{Persian (HODA)}

GraphemeNet-Persian (10 classes, $32{\times}32$) achieves 99.72\% test accuracy and 99.72\% macro-F1
with 5.34M parameters. The isotropic $3{\times}3$ scaffold captures rounded Persian loop bodies.
Training converged quickly: by epoch 2, training accuracy reached 99.49\% (val 99.53\%). Best
validation accuracy was 99.92\% at epoch 31; total wall time ${\sim}4{,}717$ seconds (larger dataset:
54,000 train + 26,000 test). Refer Table~\ref{tab:sota_consolidated} for detail.

\subsection{Japanese (Kuzushiji-49)}

GraphemeNet-Japanese (49 classes, $28{\times}28$) achieves 97.81\% test accuracy. The $3{\times}3$ scaffold captures diagonal cursive geometry more effectively than
axis-aligned asymmetric kernels. Random rotation ($\pm 8^\circ$) is enabled for pen-angle variation;
horizontal flip is disabled as mirroring any Kuzushiji glyph produces invalid characters. Training
converged at epoch 58; best validation accuracy was 97.45\%. Refer Table~\ref{tab:sota_consolidated} for detail.

\subsection{Thai (Burapha-TH)}

GraphemeNet-Thai achieves 96.93\% test accuracy and 96.93\% macro-F1 on the Burapha-TH dataset, utilizing 11.8M trainable parameters; see Section~\ref{sec:discussion}.

\subsection{English: Seven Benchmarks}
We achieved following observations, Table~\ref{tab:english_benchmarks}:
\begin{itemize}
  \item \textbf{EMNIST/Digits:} Val accuracy reached 99.02\% at epoch 1, demonstrating rapid
    convergence on the well-structured 10-class digit task.
  \item \textbf{EMNIST/Balanced:} Early stopping at epoch 37 (patience 12). Class weights [0.39,
    5.86] applied. Best val 90.21\% at epoch 25.
  \item \textbf{EMNIST/ByMerge:} Early stopping at epoch 45 (patience 12). Class weights [0.39,
    5.86]. Best val 89.48\% at epoch 18.
  \item \textbf{EMNIST/ByClass:} Early stopping at epoch 35 (patience 12). Class weights [0.29,
    5.90]---widest range due to case-sensitive 62-class split. Macro F1 of 76.20\% is substantially
    below accuracy (84.45\%), reflecting severe per-class imbalance.
  \item \textbf{PG-HWLD:} Longest training run (80 epochs, patience 20). Strongest augmentation
    ($\pm 10^\circ$, $\sigma=0.04$, dropout$=0.40$) required for 660 samples/class.
\end{itemize}

\begin{table}[h]
\centering
\caption{%
  Comparison across all non-English benchmarks; see Section \ref{sec:discussion} for discussion.
}
\label{tab:sota_consolidated}
\setlength{\tabcolsep}{3pt}
\renewcommand{\arraystretch}{1.1}
\footnotesize
\begin{tabular}{@{}p{2.2cm}p{3cm}cccp{2cm}@{}}
\toprule
\textbf{Script (Dataset)} & \textbf{Method} & \textbf{Acc.} & \textbf{F1} &
  \textbf{Params} & \textbf{Venue} \\
\midrule
 
\textbf{Devanagari} (DHCD, 46 cl.) &
  Original CNN~\cite{acharya2015}        & 98.47\% & ---     & $\sim$1M  & IEEE SKIMA 2015 \\
& BMCNNwHVCs~\cite{byerly2021}           & 99.16\% & ---     & ---       & arXiv 2021 \\
& ResNet-85~\cite{mishra2021}            & 99.72\% & ---     & $\sim$39M & IEEE INDISCON 2021 \\
& MallaNet~\cite{malla2025}              & 99.71\% & 99.71\% & 17M       & Sci.\ Rep.\ 2025 \\
\cmidrule(lr){2-6}
& \cellcolor{gray!20}\textbf{GraphemeNet-Devanagari (ours)} &
  \cellcolor{gray!20}\textbf{99.75\%} &
  \cellcolor{gray!20}\textbf{99.75\%} &
  \cellcolor{gray!20}\textbf{3.9M} &
  \cellcolor{gray!20}This work \\
\midrule
 
\textbf{Bengali} (CMATERdb 3.1.2, 50 cl.) &
  Opu et al.~\cite{opu2024}              & 97.38\% & ---     & ---       & NCA 2024 \\
& Shahariar et al.~\cite{parvez2025}     & 98.40\% & ---     & ---       & Front.\ Big Data 2025 \\
\cmidrule(lr){2-6}
& \cellcolor{gray!20}\textbf{GraphemeNet-Bengali (ours)} &
  \cellcolor{gray!20}\textbf{98.90\%} &
  \cellcolor{gray!20}\textbf{98.90\%} &
  \cellcolor{gray!20}3.9M &
  \cellcolor{gray!20}This work \\
\midrule
 
\textbf{Kannada} (Kannada-MNIST, 10 cl.) &
  Prabhu baseline~\cite{prabhu2019}      & 97.06\% & ---     & ---       & arXiv 2019 \\
\cmidrule(lr){2-6}
& \cellcolor{gray!20}\textbf{GraphemeNet-Kannada (ours)} &
  \cellcolor{gray!20}\textbf{98.29\%} &
  \cellcolor{gray!20}\textbf{98.29\%} &
  \cellcolor{gray!20}5.3M &
  \cellcolor{gray!20}This work \\
\midrule
 
\textbf{Arabic} (AHCD, 28 cl.) &
  AlShehri (DeepAHR)~\cite{alshehri2024} & 98.66\% & ---     & ---       & NCA 2024 \\
& Alkayed et al.~\cite{alkayed2024}      & 98.87\% & ---     & ---       & PCS 2024 \\
& CNN-14~\cite{bouchantouf2025}          & 99.36\% & ---     & ---       & Informatica 2025 \\
& Ali et al.\ (CNN-SVM)*~\cite{ali2023}  & 99.71\% & ---     & ---       & NCA 2023 \\
\cmidrule(lr){2-6}
& \cellcolor{gray!20}GraphemeNet-Arabic$^{\dagger\dagger}$ (ours) &
  \cellcolor{gray!20}98.10\% &
  \cellcolor{gray!20}98.09\% &
  \cellcolor{gray!20}5.4M &
  \cellcolor{gray!20}This work \\
\midrule
 
\textbf{Persian} (HODA, 10 cl.) &
  (No direct published baseline reported) & --- & --- & --- & --- \\
\cmidrule(lr){2-6}
& \cellcolor{gray!20}\textbf{GraphemeNet-Persian (ours)} &
  \cellcolor{gray!20}\textbf{99.72\%} &
  \cellcolor{gray!20}\textbf{99.72\%} &
  \cellcolor{gray!20}5.3M &
  \cellcolor{gray!20}This work \\
\midrule
 
\textbf{Japanese} (Kuzushiji-49, 49 cl.) &
  Al-Barham et al.~\cite{albarham2023}   & 95.00\% & ---     & ---         & arXiv 2023 \\
& Zhang et al.~\cite{zhang2021}          & 97.16\% & ---     & ---         & 2021 \\
\cmidrule(lr){2-6}
& \cellcolor{gray!20}\textbf{GraphemeNet-Japanese (ours)} &
  \cellcolor{gray!20}\textbf{97.81\%} &
  \cellcolor{gray!20}\textbf{97.57\%} &
  \cellcolor{gray!20}6.3M &
  \cellcolor{gray!20}This work \\
\midrule
 
\textbf{Thai} (Burapha-TH, 68 cl.) &
  Prommas et al.\ (VGGNet-19)~\cite{prommas2020} & 99.20\% & --- & ---   & ICISS 2020 \\
\cmidrule(lr){2-6}
& \cellcolor{gray!20}GraphemeNet-Thai\dag\ (ours) &
  \cellcolor{gray!20}\textbf{96.93\%} &
  \cellcolor{gray!20}\textbf{96.93\%} &
  \cellcolor{gray!20}13.4M &
  \cellcolor{gray!20}This work \\
\bottomrule
\end{tabular}
\end{table}

 
\begin{table}[htbp]
\centering
\caption{%
  GraphemeNet results on all English/Latin benchmarks with published baselines.
}
\label{tab:english_benchmarks}
\setlength{\tabcolsep}{3.5pt}
\renewcommand{\arraystretch}{1.15}
\footnotesize
\begin{tabular}{@{}p{3cm}p{2.6cm}ccccp{1.6cm}@{}}
\toprule
\textbf{Benchmark} &
  \textbf{Prior Best} &
  \textbf{Prior Acc.} &
  \textbf{GraphemeNet Acc.} &
  \textbf{GraphemeNet F1} &
  \textbf{Test Loss} &
  \textbf{Params} \\
\midrule

MNIST
  (10 cl.) &
  BMCNNwHVCs~\cite{byerly2021}$^\star$ &
  99.87\% &
  99.64\% &
  99.64\% &
  0.5113 &
  5.3M \\
\midrule

EMNIST/Digits
  (10 cl.) &
  WaveMix~\cite{jeevan2023wavemix}$^{\star\star}$ &
  99.84\% &
  99.74\% &
  99.75\% &
  --- &
  5.3M \\
\midrule

EMNIST/Letters
  (26 cl.) &
  WaveMix~\cite{jeevan2023wavemix}$^{\star\star}$ &
  95.96\% &
  94.03\% &
  94.02\% &
  0.7925 &
  5.5M \\
\midrule

EMNIST/Balanced
  (47 cl.) &
  WaveMix~\cite{jeevan2023wavemix}$^{\star\star}$ &
  91.10\% &
  \textbf{90.23\%} &
  90.14\% &
  0.9493 &
  12.1M \\
\midrule

EMNIST/ByMerge
  (47 cl., unbal.) &
  WaveMix~\cite{jeevan2023wavemix}$^{\star\star}$ &
  91.64\% &
  \textbf{89.71\%} &
  88.76\% &
  --- &
  5.7M \\
\midrule

EMNIST/ByClass
  (62 cl., unbal.) &
  WaveMix~\cite{jeevan2023wavemix}$^{\star\star}$ &
  90.87\% &
  \textbf{84.45\%} &
  76.20\% &
  0.5832 &
  21.5M \\
\midrule

PG-HWLD
  (26 cl.) &
  VGG-5~\cite{pghwld2025}$^\dagger$ &
  84.27\% &
  \textbf{96.85\%} &
  \textbf{96.89\%} &
  0.7304 &
  5.5M \\
\bottomrule
\end{tabular}
 
\medskip
\noindent\footnotesize
\textit{Note on PG-HWLD baseline.}
Szymański et al.~\cite{pghwld2025} tested four models on PG-HWLD, all
trained on the EMNIST-Letters training split: VGG-5 (84.27\%), VGG-5 with
SpinalFC (83.76\%), WaveMixLite-112/16 (83.48\%), and TextCaps (82.35\%).
VGG-5 achieves the highest reported accuracy; GraphemeNet surpasses this by
\textbf{+12.58 pp}, demonstrating substantially better generalisation to
out-of-distribution, independently collected handwriting.
\end{table}

\section{Discussion}
\label{sec:discussion}


Our results provide empirical support for a principle we term \textit{structural-prior efficiency}: for
scripts with strong geometric regularities, explicitly encoding domain-specific structural knowledge
yields accuracy gains that would otherwise require 2--10$\times$ more parameters. GraphemeNet-Devanagari
achieves 99.75\% with 3.9M parameters; the next-best result requires either 39M (ResNet-85, same
accuracy rank) or 17M (MallaNet) parameters.


The fact that one architecture with only two script-specific settings (scaffold kernel, multi-scale feature
topology) generalises across eight writing systems is non-trivial. Devanagari and Bengali share the
shirorekha ($1{\times}5$ scaffold); Kannada requires a shorter $1{\times}3$ scaffold; Arabic, Persian,
and Kuzushiji all benefit from isotropic $3{\times}3$ scaffolds for different geometric reasons; Thai and
English share the dual-axis stem. The LCR and gated decoder are fully shared across all scripts.

Thai is the one script where the encoder scaffold is intentionally removed: the complex embedded
circular loops of Thai consonants are better handled by a richer triple-path stem and an unconstrained
end-to-end encoder than by a scalar scaffold bypass that would blur loop-body geometry across
encoder stages.

GraphemeNet-Thai model, evaluated on the Burapha-TH dataset, is loaded from a folder-per-class structure using a TensorFlow pipeline (with PyTorch compatibility). The dataset, extracted from thai-dataset.zip, consists of 68 Thai character classes (e.g., KO KAI, KHO KHAI) with 56,995 training, 6,332 validation, and 13,600 test images. The GraphemeNet-Thai model, with 11.8M trainable parameters, is trained for 100 epochs (early stopping at Epoch 53) on a CUDA-enabled GPU, achieving a best validation accuracy of 96.62\% and a final test accuracy of 96.93\%, with a Macro F1-score of 96.92\% and a test loss of 0.831. 


GraphemeNet-Arabic achieves 98.10\%, below the 99.36\% and 99.71\% reported by CNN-14 and the
CNN-SVM hybrid respectively. These higher-performing baselines use significantly larger models or
two-stage pipelines; GraphemeNet-Arabic achieves competitive single-model performance within the unified
architecture framework. Closing this gap is an objective for future work.


EMNIST/ByClass is the most challenging benchmark (62 classes, full case-sensitive, unbalanced). The
Macro F1 of 76.20\% is substantially below accuracy (84.45\%), reflecting the difficulty of distinguishing
visually similar upper/lower case pairs in an unbalanced setting with class weight ranges reaching
[0.29, 5.90]. EMNIST/Balanced (47 classes, class-balanced) reaches 90.23\%, confirming that class
imbalance is the primary challenge. Macro F1 gap for ByClass (76.20\% vs 84.45\% accuracy) is a practically important finding:
accuracy overstates recognition capability when classes are imbalanced. Future work should report
both metrics for all imbalanced splits.


All Indic, Semitic, Japanese (TF attempt), and English (EMNIST, MNIST, PG-HWLD) models use
TensorFlow 2.x/Keras/PyTorch. The final Japanese Kuzushiji model, all three unbalanced
EMNIST splits (Balanced, ByMerge, ByClass), and Thai were implemented in PyTorch (Method 2)
due to superior training stability and control over weighted sampling. The Thai TF implementation
(Method 1) is also provided and uses the same architecture with early stopping patience 15 and
50 training epochs. Both frameworks yield comparable results for the same architecture.



 
Together, these claims constitute an empirical test of
\emph{structural-prior efficiency}: the productive question for multi-script HCR
is not \textit{how many parameters does it take to learn this script?} but
\textit{what structural knowledge of this script can be encoded directly, and
how much does that encoding reduce the learning burden?}

\section{Conclusion}

GraphemeNet achieves a new state-of-the-art of 99.75\% on DHCD with 3.9M parameters ($10\times$ more
efficient than ResNet-85, $4.4\times$ more efficient than MallaNet), confirmed by confusion matrix
analysis showing Macro Precision/Recall/F1 all at 99.8\%. Across fourteen additional benchmarks the
architecture demonstrates strong and consistent generalisation: EMNIST/Digits 99.74\%, Persian
99.72\%, MNIST 99.64\%, Bengali 98.90\%, Kannada 98.29\%, Arabic 98.10\%, Kuzushiji-49 97.81\%,
PG-HWLD 96.85\%, EMNIST/Letters 94.03\%, EMNIST/Balanced 90.23\%, EMNIST/ByMerge 89.71\%,
and EMNIST/ByClass 84.45\%.

Ablation experiments on DHCD confirm that Persistent Scaffold Injection is the single most
important architectural component: removing the scaffold drops test accuracy by 0.15\% (99.75\% $\to$
99.60\%) while reducing parameters by only 18K, demonstrating that the scaffold's value is
informational rather than parametric. An augmented STM variant (4.2M parameters, 3-way gated
decoder) achieves 99.72\%, confirming that additional decoder complexity does not improve on the
base architecture for this benchmark.

Our central finding---that explicit script-structural priors encoded architecturally can simultaneously
improve accuracy and reduce parameter requirements---holds across all eight writing systems
evaluated, establishing \textit{structural-prior efficiency} as a broadly applicable principle in
multi-script HCR. The Thai variant (GraphemeNet-Thai) demonstrates a complementary insight: when loop
geometry is too complex for a scalar scaffold bypass, a richer multi-path stem with an unconstrained
encoder is the appropriate prior.

The productive question for future HCR research is not \textit{how many parameters does it take to
learn this script?} but rather \textit{what structural knowledge of this script can we encode directly,
and how much does that encoding reduce the learning burden?} For scripts with strong geometric
regularities---which includes virtually all writing systems---the answer is: considerably. Extend
coverage to additional Indic scripts, Latin scripts, and other scripts.

{
\small
\bibliographystyle{unsrt}

}

\newpage
\appendix
\section{Dataset Sample Images}
\label{app:datasets}
Dataset samples illustrate the visual diversity and intra-class variation encountered during training.



\begin{figure}[htbp]
    \centering
    \includegraphics[width=0.6\linewidth]{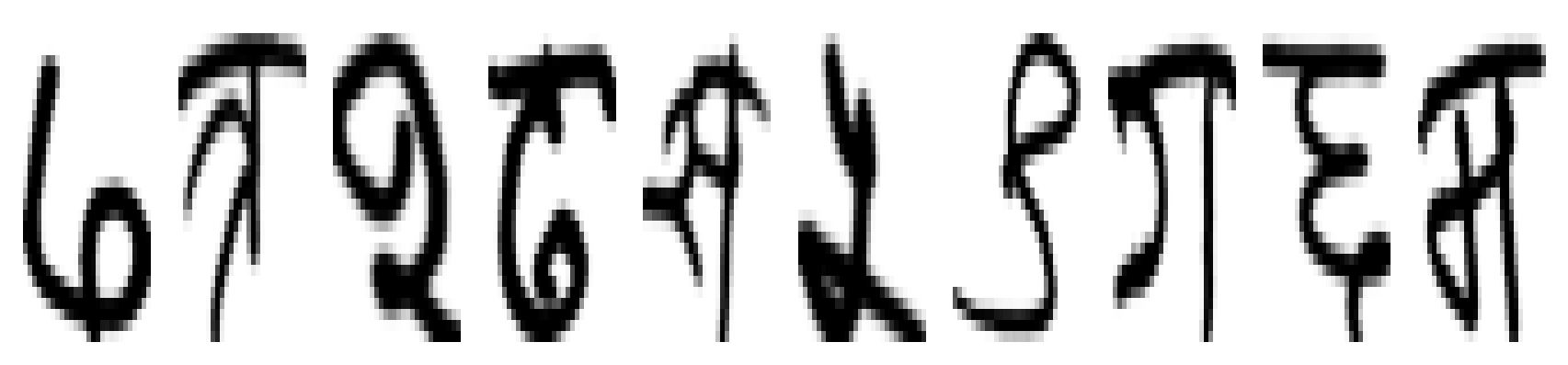}
    \caption{Sample images from the Devanagari DHCD dataset (46 classes, 32$\times$32). Each row corresponds to one character class; columns show intra-class writer variation.}
    \label{fig:samples_devanagari}
\end{figure}

\begin{figure}[h]
    \centering
    \includegraphics[width=0.6\linewidth]{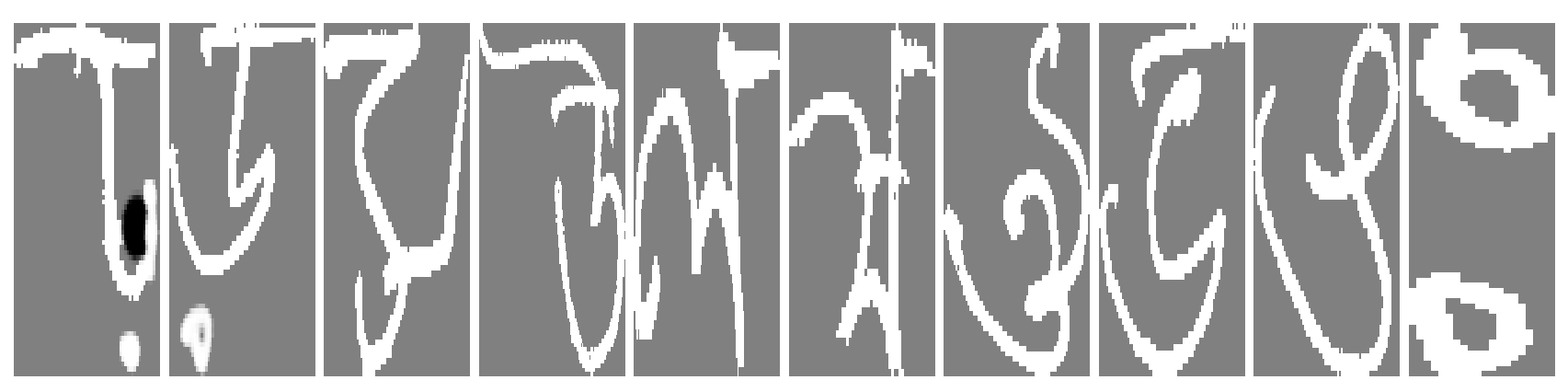}
    \caption{Sample images from the Bengali CMATERdb~3.1.2 dataset (50 classes, 32$\times$32).}
    \label{fig:samples_bengali}
\end{figure}

\begin{figure}[h]
    \centering
    \includegraphics[width=0.6\linewidth]{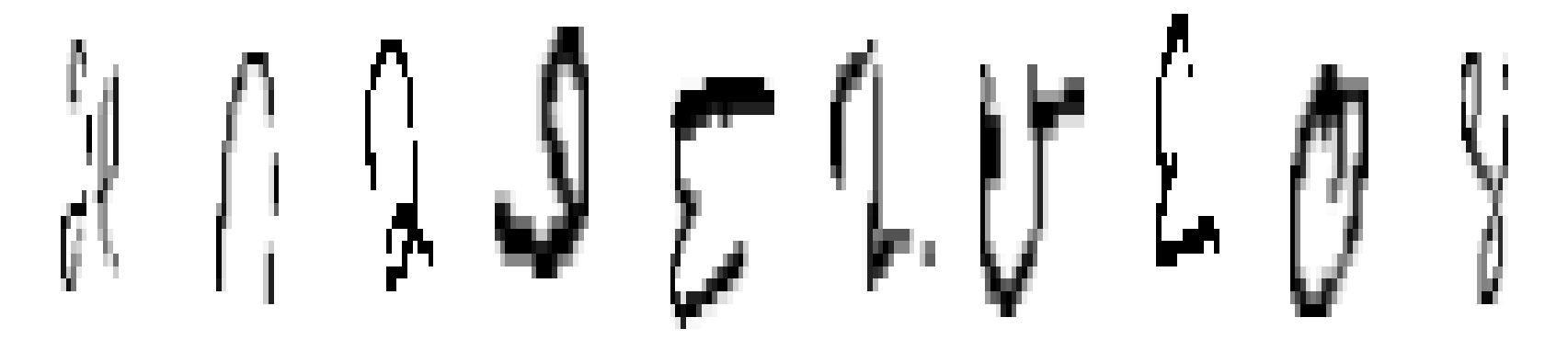}
    \caption{Sample images from the Kannada-MNIST dataset (10 classes, 28$\times$28). Characters are defined by closed loops and rounded curves without a shirorekha headline.}
    \label{fig:samples_kannada}
\end{figure}

\begin{figure}[!htbp]
    \centering
    \includegraphics[width=0.6\linewidth]{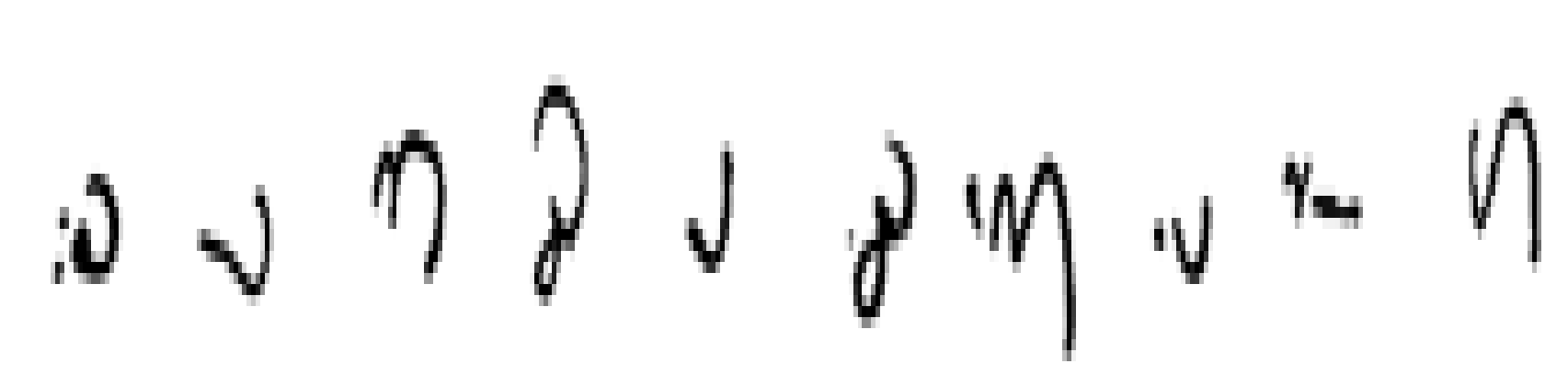}
    \caption{Sample images from the Arabic AHCD dataset (28 classes, 32$\times$32). Letter identity depends jointly on base-stroke shape and dot diacritic count and position.}
    \label{fig:samples_arabic}
\end{figure}

\begin{figure}[!htbp]
    \centering
    \includegraphics[width=0.6\linewidth]{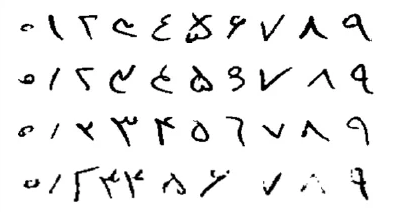}
    \caption{Sample images from the Persian HODA dataset (10 digit classes, 32$\times$32).}
    \label{fig:samples_persian}
\end{figure}


\begin{figure}[!htbp]
    \centering
    \includegraphics[width=0.6\linewidth]{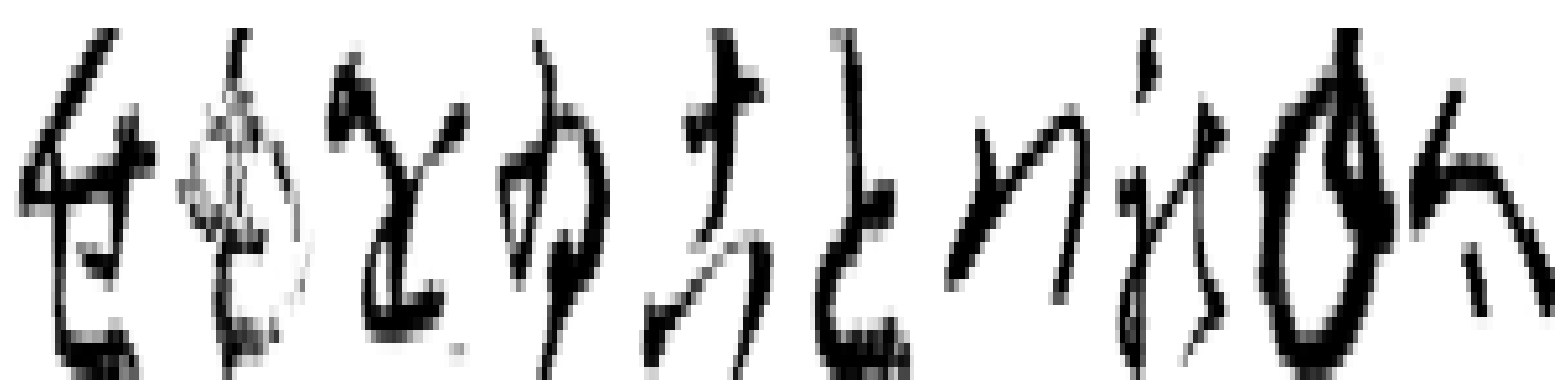}
    \caption{Sample images from the Kuzushiji-49 dataset (49 Hiragana classes, 28$\times$28) in classical cursive style. Flowing diagonal strokes with harai tails whose angular variation encodes character class.}
    \label{fig:samples_kuzushiji}
\end{figure}

\begin{figure}[!htbp]
    \centering
    \includegraphics[width=0.6\linewidth]{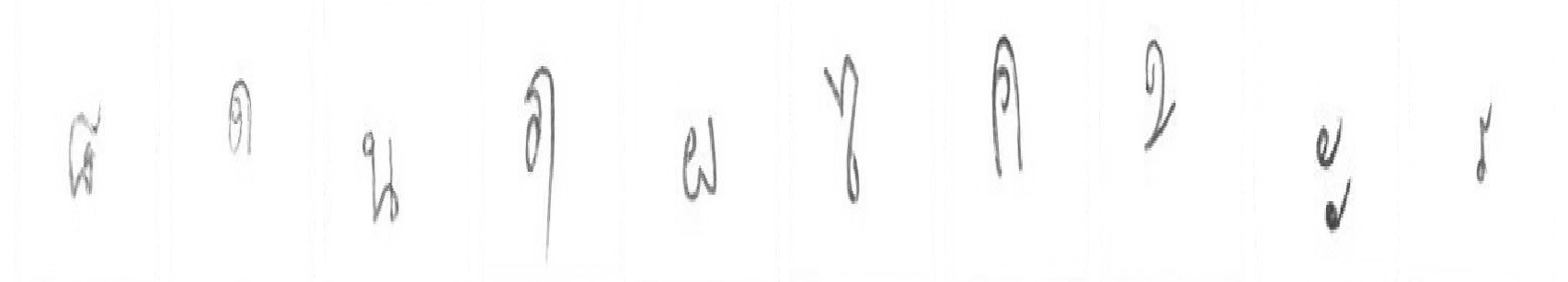}
    \caption{Sample images from the Burapha-TH Thai dataset (68 classes, 64$\times$64). Thai consonants are characterised by complex embedded circular loops and hovering diacritics at varying heights above the baseline.}
    \label{fig:samples_thai}
\end{figure}

\begin{figure}[!htbp]
    \centering
    \includegraphics[width=0.4\linewidth]{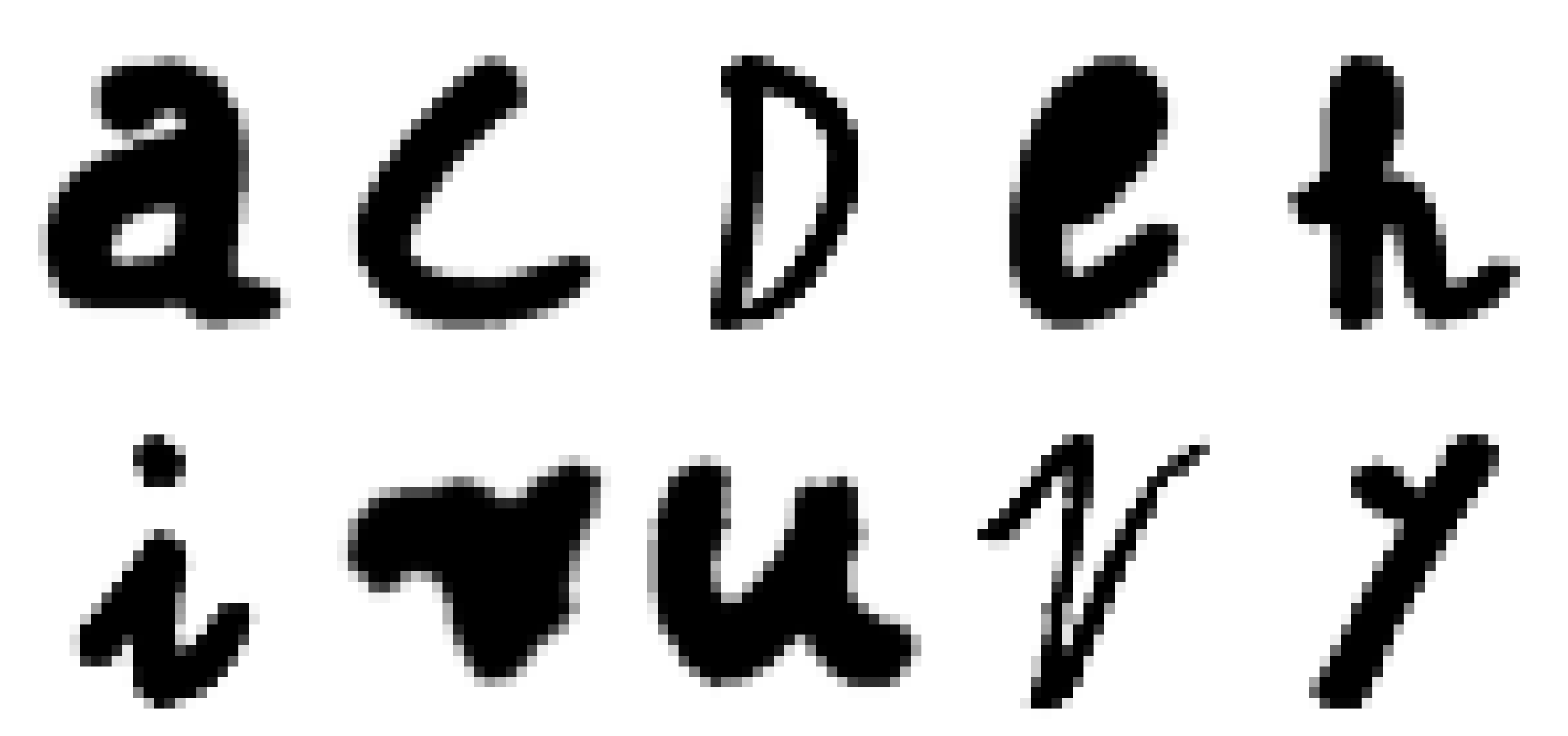}
    \caption{Sample images from PG-HWLD (26 letter classes, 28$\times$28), an independently collected handwriting dataset distinct from the EMNIST distribution.}
    \label{fig:samples_pghwld}
\end{figure}

\begin{figure}[htbp]
    \centering
    \includegraphics[width=0.5\linewidth]{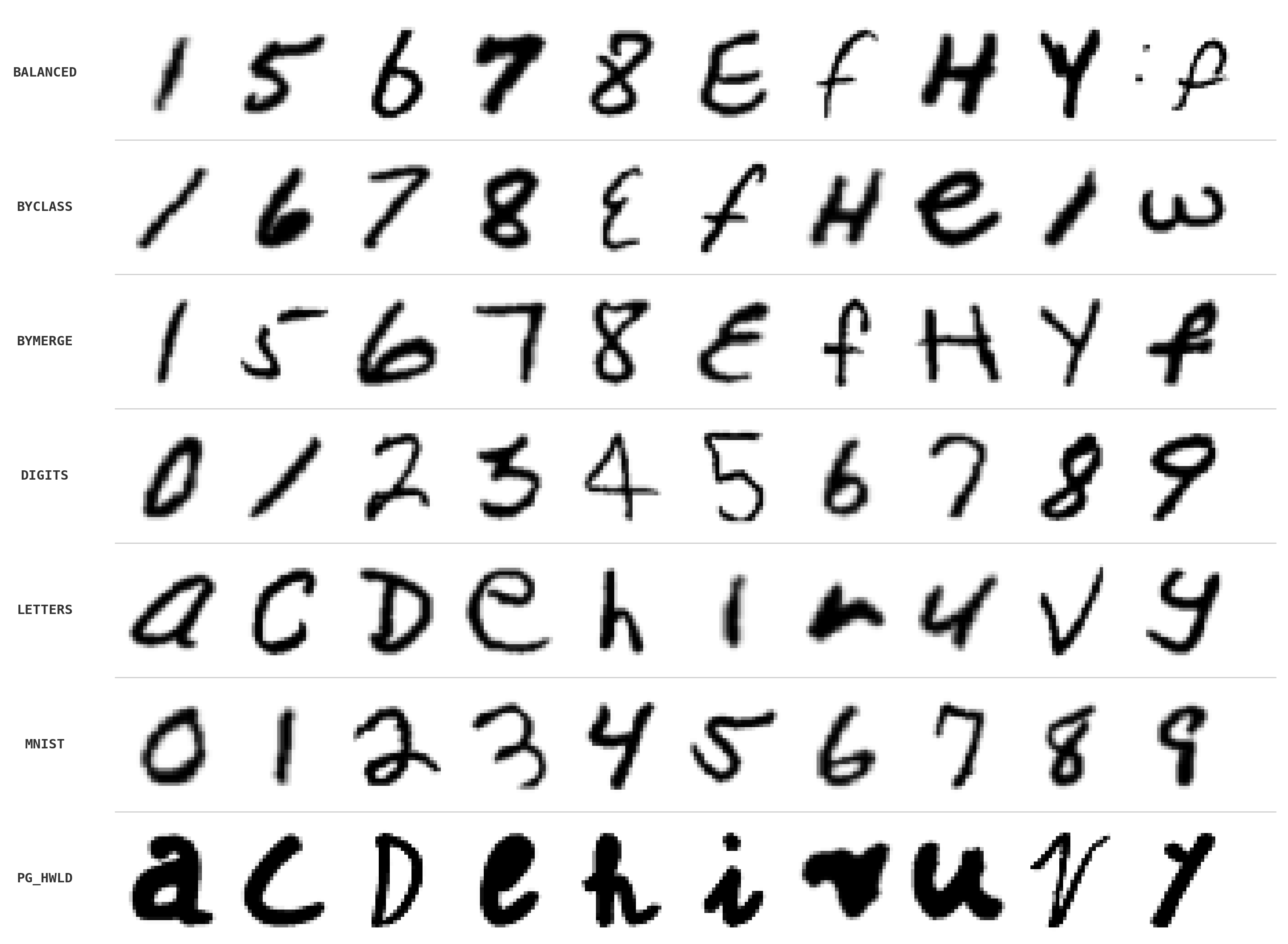}
    \caption{Sample images from the EMNIST benchmark suite (28$\times$28). From top to bottom: Letters (26 cl.), Digits (10 cl.), Balanced (47 cl.), ByMerge (47 cl.), ByClass (62 cl.). ByClass is case-sensitive and heavily unbalanced, with class weights ranging from 0.29 to 5.90.}
    \label{fig:samples_emnist}
\end{figure}

\begin{figure}[htbp]
    \centering
    \includegraphics[width=0.4\linewidth]{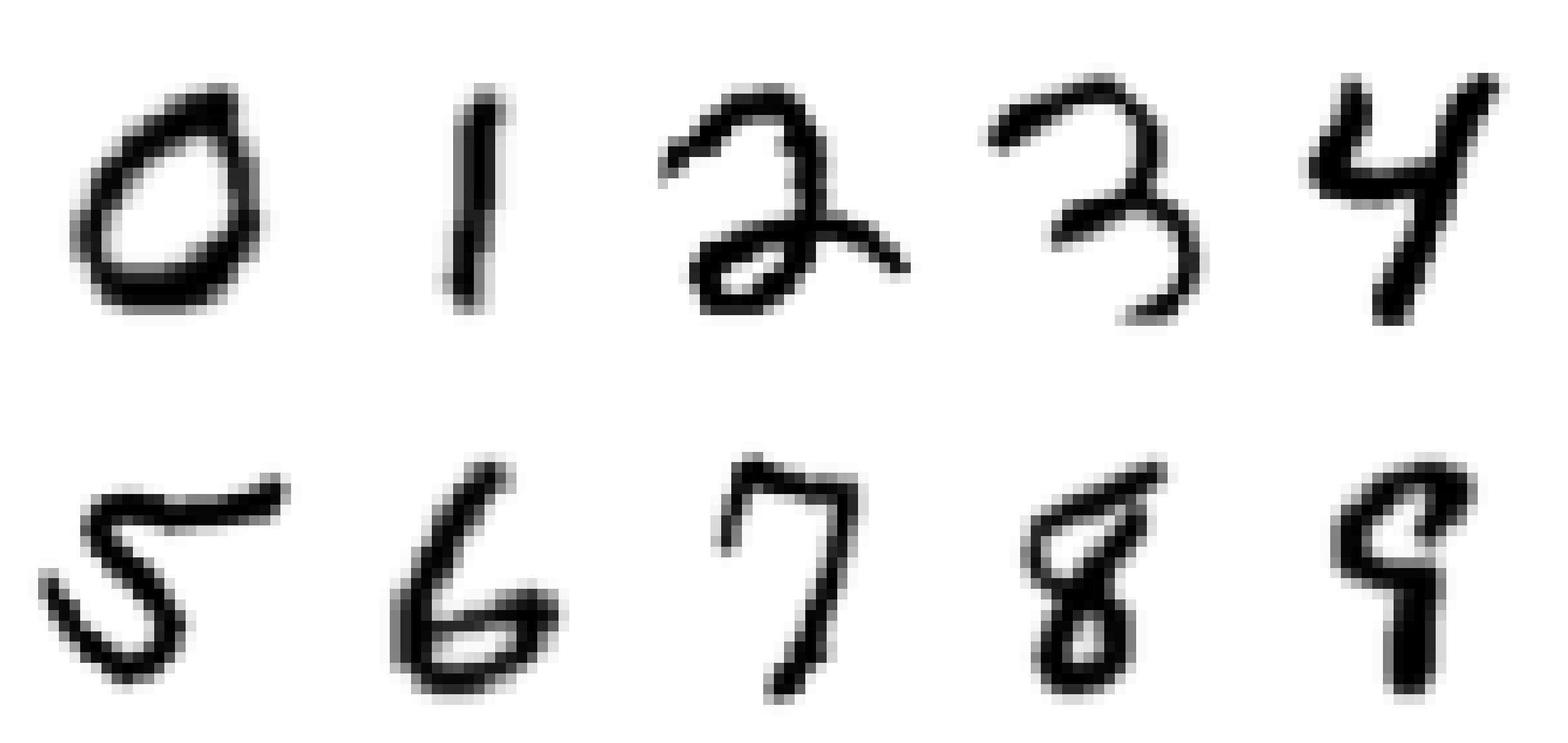}
    \caption{Sample images from MNIST (10 digit classes, 28$\times$28).}
    \label{fig:samples_mnist}
\end{figure}
\clearpage
\newpage
\end{document}